\documentclass[final]{cvpr}

\usepackage{times}
\usepackage{epsfig}
\usepackage{graphicx}
\usepackage{amsmath}
\usepackage{amssymb}

\usepackage[pagebackref=true,breaklinks=true,colorlinks,bookmarks=false]{hyperref}

\def\cvprPaperID{6278} 
\def\confYear{CVPR 2021}
\begin{document}

\title{Toward Accurate and Realistic Outfits Visualization with Attention to Details}

\author{
Kedan Li$^{1,2}$\\
\and
Min Jin Chong$^{1,2}$\\
\and
Jeffrey Zhang$^{1,2}$\\
\and
Jingen Liu$^3$\\
\and
{\tt\small \{kedan, minjin, jeff\}@revery.ai, jingenliu@gmail.com}\\
\small$^1$University of Illinois, Urbana Champaign.
\small$^2$Revery AI Inc.
\small$^3$JD AI Research.
}

\maketitle

\thispagestyle{empty}
\begin{abstract}
Virtual try-on methods aim to generate images of fashion models wearing arbitrary combinations of garments. 
This is a challenging task because the generated image must appear realistic and accurately display the interaction between garments. 
Prior works produce images that are filled with artifacts and fail to capture important visual details necessary for commercial applications.
We propose Outfit Visualization Net (OVNet) to capture these important details (e.g. buttons, shading, textures, realistic hemlines, and interactions between garments) and produce high quality multiple-garment virtual try-on images.
OVNet consists of 1) a semantic layout generator and 2) an image generation pipeline using multiple coordinated warps. 
We train the warper to output multiple warps using a cascade loss, which refines each successive warp to focus on poorly generated regions of a previous warp and yields consistent improvements in detail.
In addition, we introduce a method for matching outfits with the most suitable model and produce significant improvements for both our and other previous try-on methods. 
Through quantitative and qualitative analysis, we demonstrate our method generates substantially higher-quality studio images compared to prior works for multi-garment outfits.
An interactive interface powered by this method has been deployed on fashion e-commerce websites and received overwhelmingly positive feedback.
\end{abstract}

 \begin{figure}
\begin{center}
   \includegraphics[width=0.99\linewidth]{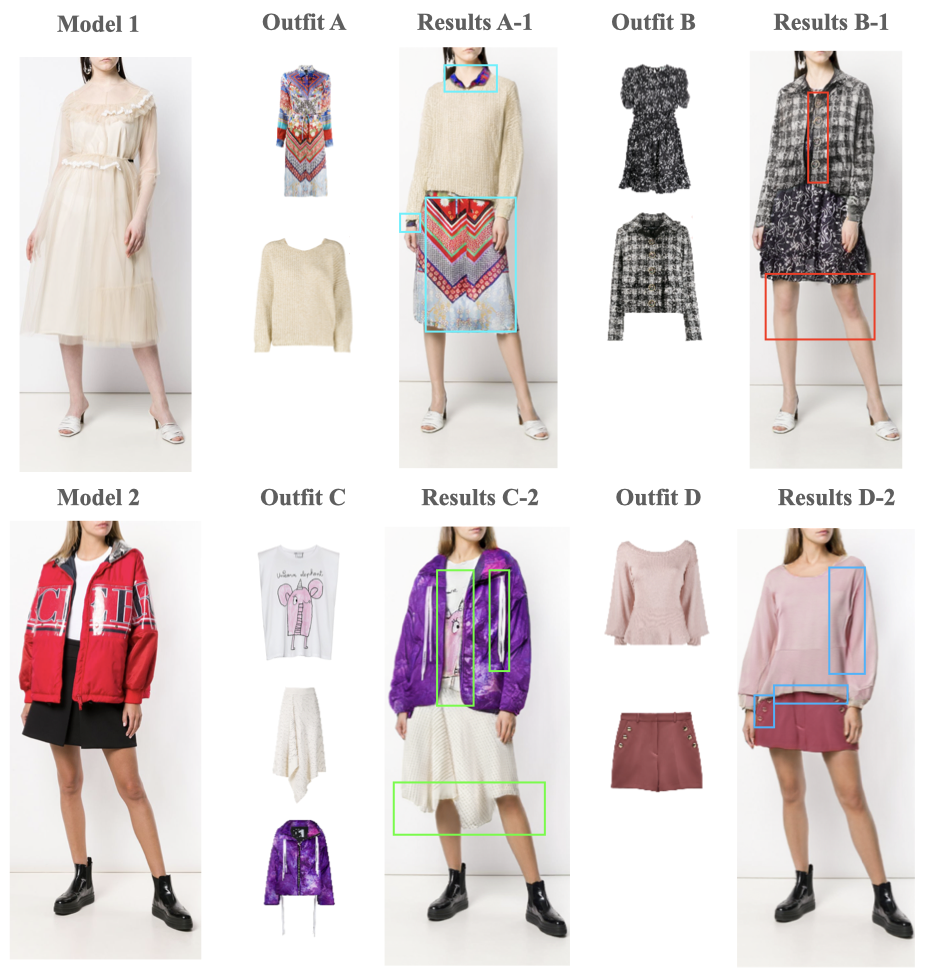}
\end{center}
           \vspace{-0.5cm}  
    \caption{Our method takes in a model image and multiple neutral garments images as inputs, and generates a high quality image of the selected model wearing the garments. {\bf Pay careful attention to details} of the garment properties that are accurately portrayed (e.g., the patterns on the dress (A-1), the unicorn and the string (C-2), the hemline (C-2), buttons (B-1, D-2), and the lengths of the garments); the interaction between multiple garments has been captured (e.g., the collar and sleeve coming out of the sweater (A-1), the open outerwear cast shading (B-1, C-2) to the garment beneath); the interaction between the garment and the person is natural (e.g., the loose sleeves, the folds by the arm (D-2), and the shadows casted on the leg by the dresses); and skin is generated realistically (B-1). See image without bounding box in Appendix.}  
       \vspace{-0.5cm}  
          \label{fig:header}
\end{figure}

\section{Introduction}\label{sec:intro}

While e-commerce has brought convenience to many aspects of our lives, shopping online is difficult for fashion consumers who want to try-on garments and outfits before deciding to buy them~\cite{Vaccaro:2018:DFP:3173574.3174201}. In most online shopping experiences, we are only given a neutral product image of a garment or a single example of a model wearing the garment, and users have to imagine how the garment would look in different settings (e.g. with different garments, on different models etc.). As a result, there has been a considerable amount of literature on synthesizing people wearing garments~\cite{han2017viton,wang2018toward,Rocco17,Dong2018SoftGatedWF,Han_2019_ICCV,tprvton,Neuberger_2020_CVPR,Jandial2020SieveNetAU,Issenhuth2020DoNM}. 

Three natural cases arise when shopping online. A user may want to see {\bf(a) any image of a model} wearing a chosen set of garments (outfit) to visualize a combination; {\bf(b) any image of themselves} wearing the outfit to see how the garments interact; and {\bf(c) an image of themselves} wearing the outfit (the VITON case~\cite{han2017viton,wang2018toward,Rocco17,Dong2018SoftGatedWF,Han_2019_ICCV,tprvton,Jandial2020SieveNetAU,Issenhuth2020DoNM}). In all cases, users expect the image to capture the visual features of the garments and the physical interactions between them. However, current methods have problems capturing details of shading, texture, drape and folds. Getting these right is crucial for shoppers to make purchase decisions. 

In this work, we introduce a variety of innovations that substantially improve upon the synthesis of details (Figure~\ref{fig:header}). Our proposed method not only produces accurate textures, necklines, and hemlines, but also can drape multiple garments with realistic overlay and shading. The drape can adapt to the body pose and generate natural creases, folds, and shading. Skin and background are also generated, with appropriate shadows casted from the garments (Figure~\ref{fig:header}). Our method significantly outperforms prior work in multi-garment image synthesis as shown in Figure~\ref{fig:amazon_compare}.

While other virtual try-on (VITON) methods~\cite{han2017viton,wang2018toward,Rocco17,Dong2018SoftGatedWF,Han_2019_ICCV,tprvton,Issenhuth2020DoNM} focused on {\bf single} garment try-on, Neuberger \etal proposed O-VITON~\cite{Neuberger_2020_CVPR}, which transfers {\bf multiple} garments from model to model. In comparison, our system takes garments from neutral garment photographs and transfers them to a model. This distinction is commercially important because it is easier and cheaper to obtain neutral pictures. The formatting is also consistent across different sites, meaning no extra work is required for the merchants. Also, O-VITON~\cite{Neuberger_2020_CVPR} encodes garments into {\bf feature vectors} and broadcasts the vectors onto a layout to produce the image. Such a formulation can handle complex garment shapes (a major difficulty for multi-garment try-on) but results in a loss of spatial patterns (e.g., logos, prints, buttons), making it hard to synthesize texture details accurately. 
In contrast, other VITON literature~\cite{han2017viton,wang2018toward,Han_2019_ICCV,tprvton,Issenhuth2020DoNM} uses {\bf warping}, which faithfully perseveres details. However, they only demonstrate success with warping single garments of simple shapes (mostly). Warping multiple garments with complicated shapes has not yet been achieved.

\begin{figure}
\begin{center}
\vspace{-0.2cm}  
	\includegraphics[width=0.99\linewidth]{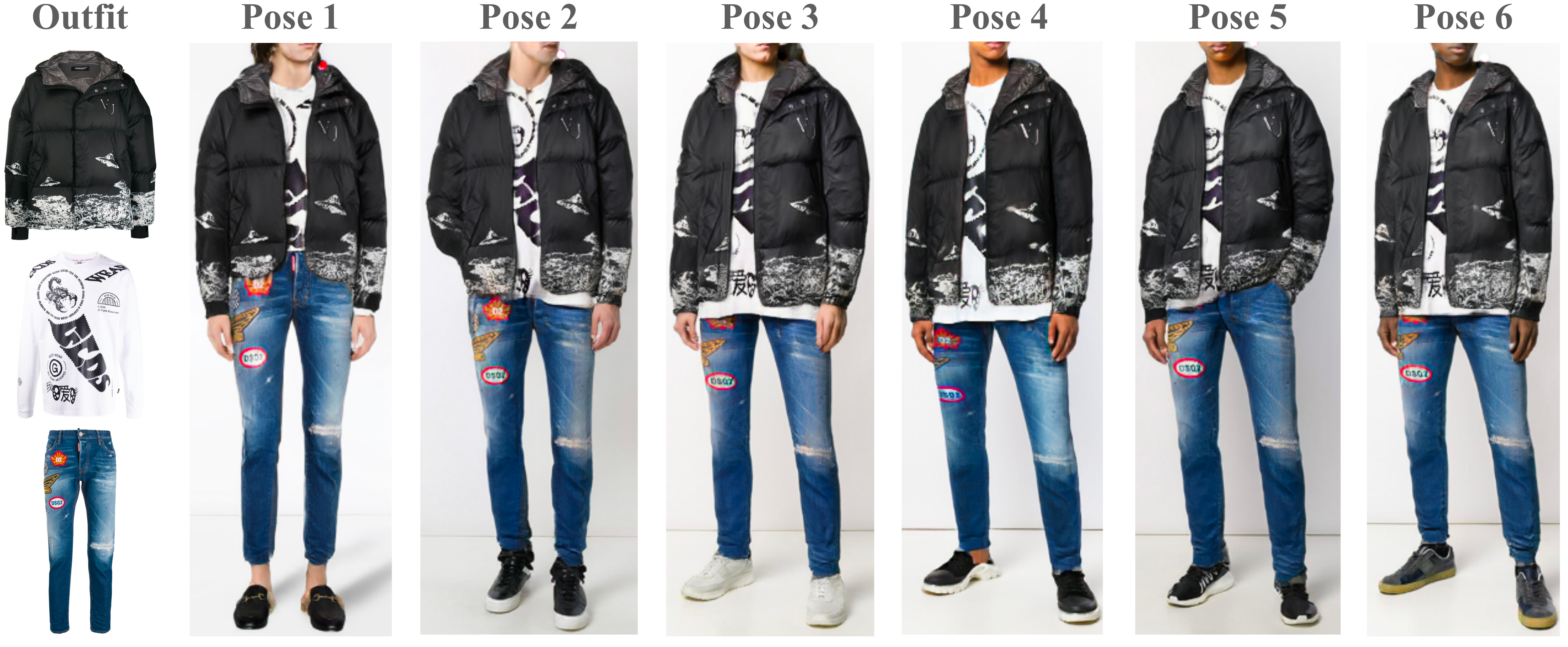}
    \caption{We show a sequence of visualizations for the same outfit generated on different reference models. Our generation method is able to adapt to a diverse set of poses, skin-tones, and hand positions. When the hand is in the pocket, the jeans plump up and connect seamlessly with the jacket (Pose 2 \& 5).}    
    \label{fig:multiple_models}
    \end{center}
\vspace{-0.8cm}  
\end{figure}

In this work, we directly address the challenge of warping multiple garments, while also being able to accurately transfer textures between complicated garment shapes (Figure~\ref{fig:header}). Our procedure uses multiple warps, which can handle (say) open jackets, and can generate buttons, zippers, logos, and collars correctly (Figure~\ref{fig:multiple_models}). The warpers are trained end-to-end with the generator and learn to coordinate through a cascading loss, which encourages subsequent warps to address errors made by earlier warps.
Using multiple coordinated warps produces substantial quantitative and qualitative improvements over prior single-warp methods~\cite{han2017viton,wang2018toward,Dong2018SoftGatedWF,Han_2019_ICCV,tprvton,Jandial2020SieveNetAU}.

Finally, because publicly available try-on datasets do not contain rich garment categories, we test on a dataset with all available garment categories from multiple fashion e-commerce websites. Evaluation on this new dataset shows that using multiple warps consistently outperforms single warp baselines in this new setting, demonstrated both quantitatively (Table~\ref{table:multi_cat}) and qualitatively (Figure~\ref{fig:multi-warp-qualitative}). Our try-on system also produces higher quality images compared to prior works on both single and multi-garment generation (Table~\ref{table:is_ssim} and~\ref{table:fid}, and Figure~\ref{fig:amazon_compare}). Furthermore, we introduce a procedure for matching garment-pose pairs, which yields significant improvement for both our and previous image generation pipelines in scenarios {\bf (a) and (b)} (Table~\ref{table:fid}). Lastly, we conduct a user study comparing our generated images with real commercial photos, simulating the effectiveness of e-commerce sites replacing real photographs of models with our synthesized images. Results show over 50\% of our synthesized images were thought to be real even with references to real images (Table~\ref{table:user_study}). Furthermore, our method is fast enough to integrate with interactive user-interfaces, where users can select garments and see generated visualizations in real-time. A live demo of an virtual try-on shopping interface powered by our method is publicly available \footnote[10]{https://demo.revery.ai}.

As a summary of our contributions:
\begin{itemize}
	\item We introduce OVNet - the first multi-garment try-on framework that generates high quality images at latencies low enough to integrate with interactive software. 
	\item We are the first warping-based try-on method that supports multi-garment synthesis on all garment types.
	\item We introduce a garment-pose matching procedure that significantly enhances our method and prior methods.
	\item Our results strongly outperform prior works, both quantitatively and qualitatively.
	\item We evaluate on a dataset with all available garment categories from multiple fashion e-commerce sites, and show that our method works with all categories.
\end{itemize}

 \begin{figure}
\begin{center}
   \includegraphics[width=1\linewidth]{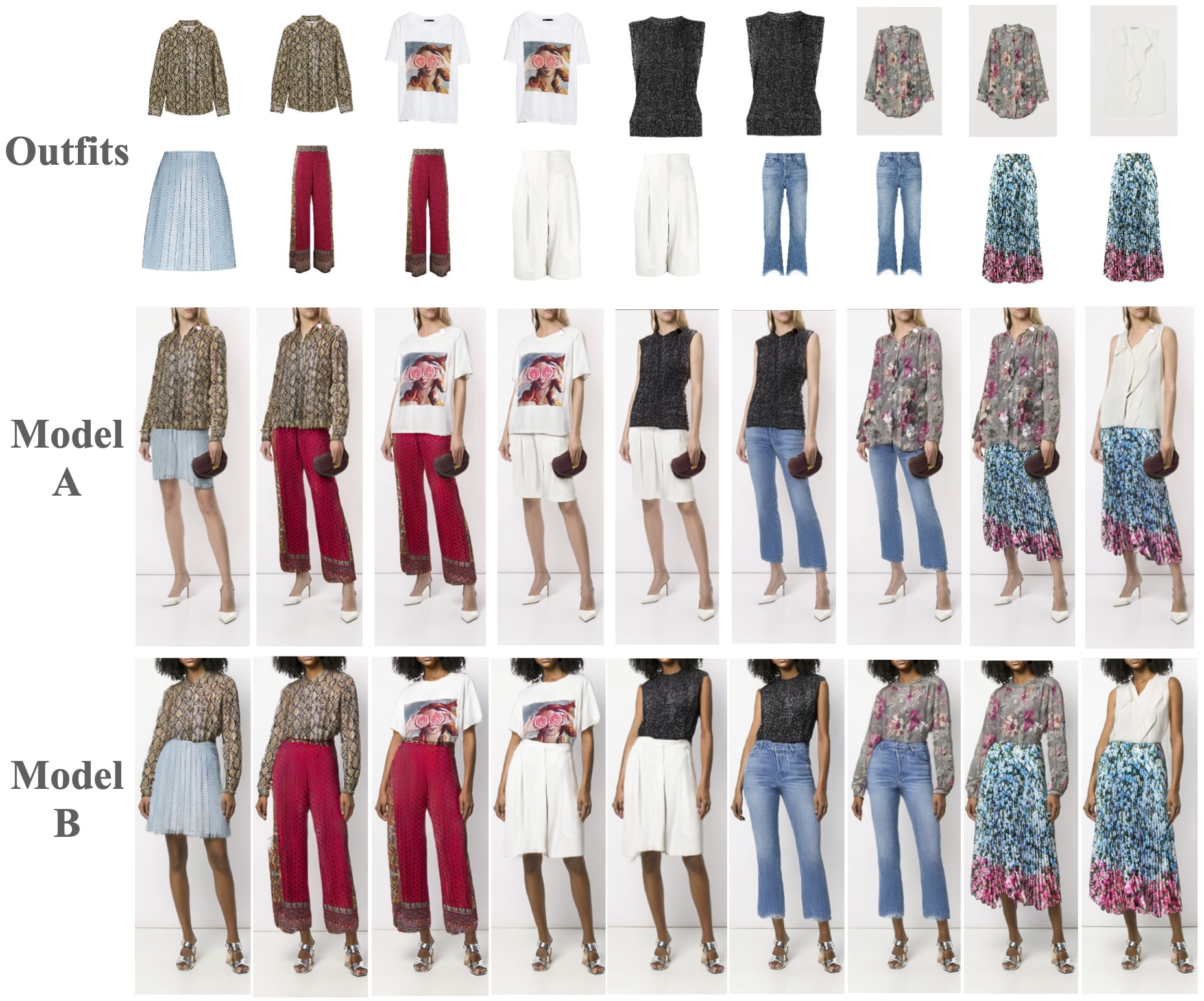}
\end{center}
           \vspace{-0.3cm}  
   \caption{ The figure shows a sequence of outfit visualizations produced by our method on two different models. Our method can modify one garment at a time, leaving the rest of the image untouched. The details of the garments' shape are accurately represented (e.g., neckline shape, skirt length, pant width, etc.) and consistent on both models. The garment interactions of the top (hanging or tucked-in) also vary between poses.
}
       \vspace{-0.5cm}  

   \label{fig:outfit-visualization}
\end{figure}

\section{Related Work}
There are multiple ways to tackle virtual try-on. One solution is to use 3D modeling and rendering~\cite{Deep3DPose,drape,patel20tailornet}, but obtaining 3D measurements of the garments and users is difficult and costly. A more economic and scalable approach is to synthesize images without 3D measurements. We discuss the variations of this approach in detail. 

{\bf Image synthesis:} Spatial transformer networks estimate geometric transformations using neural networks~\cite{NIPS2015_5854}. 
Subsequent work~\cite{warpnet,Rocco17} shows how to warp one object onto another. Warping works with images of rigid objects~\cite{Ji2017DeepVM,lin2018stgan} and non-rigid objects (e.g., clothing)~\cite{han2017viton,Dong2019TowardsMG,wang2018toward}. 

In contrast to using a single warp with high degree of freedom, our work coordinates multiple spatial warps to support garments of complex shape. We use U-Net to combine multiple warps into a single image. U-Net is commonly used for inpainting methods, which tackle filling in missing portions of an image~\cite{Yang_2017_CVPR,Liu_2018_ECCV,Yu2018GenerativeII,Yu_2019_ICCV}). Han \etal~\cite{Han2019CompatibleAD,inpainting-based} also show inpainting methods can complete missing clothing items on people. 

{\bf Generating clothed people:} Zhu \etal~\cite{zhu2017be} uses a conditional GAN to generate images based on pose skeletons and text descriptions of garments. SwapNet~\cite{Raj2018SwapNetIB} learns to transfer clothes from person A to person B by disentangling clothing and pose features. Hsiao \etal~\cite{hsiao2019fashionplus} learns a fashion model synthesis network using per-garment encodings to enable minimal edits to specific items. Recently, Men~\etal~\cite{men2020controllable} proposed a person image synthesis method, controllable through interpolating style and pose representations. These methods use feature vectors as visual representations, and thus cannot preserve geometric patterns (e.g, logo, prints). Our method warps garments and directly uses the warped images to generate our result.

{\bf Garment \& body matching} underlie our method to match garments to models. Tsiao \etal~\cite{Hsiao2019DressingFD} learns a shape embedding to enable matching between human bodies and well-fitting clothing items. 
Prior work estimates the shape of the human body~\cite{Bogo:ECCV:2016,Kanazawa2017EndtoEndRO}, clothing items~\cite{Danerek2017DeepGarment3,Jeong2015GarmentCF} and both~\cite{Natsume2019SiCloPeS,Saito2019PIFuPI}, through 2D images. The DensePose~\cite{Guler_2018_CVPR} descriptor helps model the deformation and shading of clothes and has been adopted by recent work~\cite{Neverova2018DensePT,Grigorev2019CoordinateBasedTI,Wu2018M2ETryON,8836494,Chen_2019_ICCV_Workshops,inpainting-based}. 

{\bf Virtual try-on} (VITON) maps a single garment onto a model image. VITON~\cite{han2017viton} first proposed using TPS transformation to create a warp, followed by a generation network to synthesize the final output. CP-VTON~\cite{wang2018toward} improves this method by using a differentiable component for TPS transformation. Han~\etal~\cite{Han_2019_ICCV} uses a flow estimation network to enable more degrees of freedom for the warp. Issenhuth~\etal~\cite{Issenhuth2020DoNM} proposed a teacher-student training paradigm to warp without relying on masks. To enable shape changes (e.g., short sleeve to long sleeve), a common procedure has been to predict a semantic layout of body segments and clothes to assist with image generation~\cite{tprvton,Jandial2020SieveNetAU,YuVTNFPAI,Raffiee2020GarmentGANPA, Han_2019_ICCV}. More recent works proposed architectural improvements toward better preservation of details ~\cite{Wang2019DownTT,Raffiee2020GarmentGANPA} and adding adversarial training during the refinement phase to improve image realism~\cite{Dong2018SoftGatedWF, YuVTNFPAI,tprvton,Raffiee2020GarmentGANPA}. Others followed similar procedures~\cite{Song2019SPVITONSI,Lee_2019_ICCV_Workshops,Ayush_2019_ICCV_Workshops}. The virtual try-on task has also been extended to multi-view scenarios and videos~\cite{Dong2019TowardsMG,Dong_2019_ICCV}. In summary, recent work in VITON managed to preserve garment details, but only for {\bf single garment}, with {\bf simple shapes} (mostly tops).

{\bf Outfit try-on:} Neuberger~\etal~\cite{Neuberger_2020_CVPR} proposed a virtual try-on method that works for multiple garments. The method relies on visual feature vector encoding rather than warping, which falls short in preserving textures comparing to other VITON methods. To make up for deficiencies, they proposed an online optimization step that requires fine-tuning a generator using a discriminator for every query. Performing such an operation is massively expensive (requires multiple rounds of gradient computation and back-propagation), making it unrealistic to respond to user queries in real-time. In comparison, our method produces images of significantly better quality (Figure~\ref{fig:amazon_compare}) and requires much less computation ($<$2s latency on a K80). 
\begin{figure*}
\begin{center}
   \includegraphics[width=0.8\linewidth]{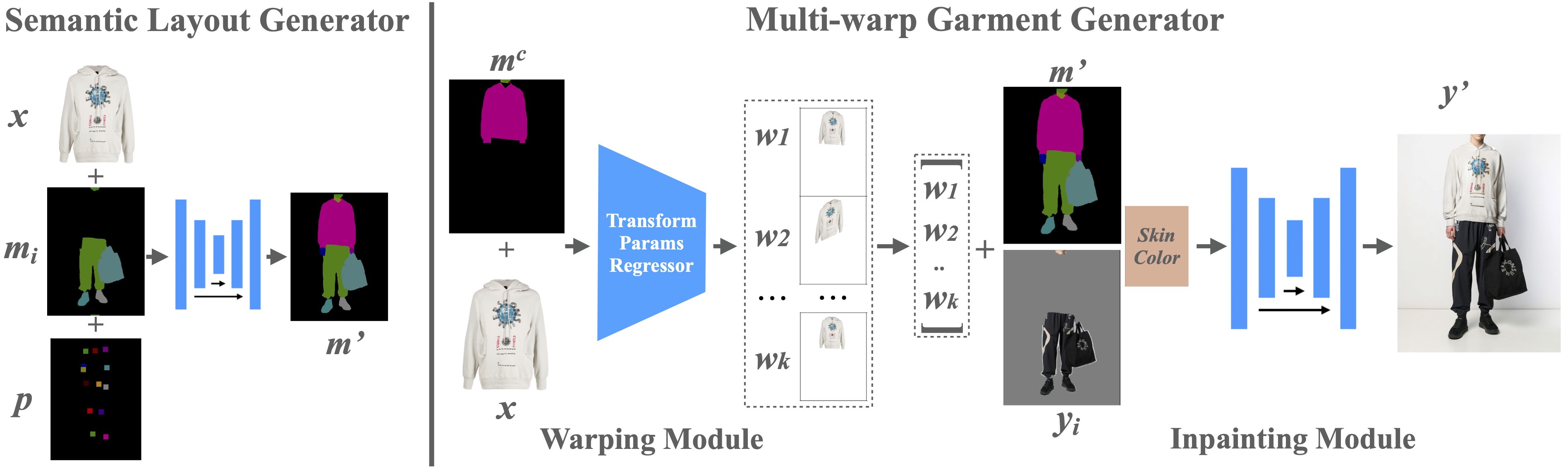}
\end{center}
           \vspace{-0.35cm}  

   \caption{{\bf Outfit Visualization Net}, which synthesizes an image of a model wearing multiple garments, consists of two components. The Semantic Layout Generator $G_{layout}$ (left) takes in the garment image $x$, the pose representation $p$ and an incomplete semantic layout $m_i$, and learns to reconstruct the ground truth layout $m$. The multi-warp garment generator $G_{garment}$ (right) has two modules. The {\bf warping module} is a spatial transformer that takes in the garment image $x$ and its semantic layout $m^c$ and regress $k$ sets of transformation parameters ${\theta_1..\theta_k}$. It then samples $k$ warps $w_1..w_k$ where $ w_1=\mathcal{W}(x,\theta_1)$, using the predicted transformations. The {\bf inpainting module} takes in the predicted warps ${w_1..w_k}$, the full semantic layout $m'$, the skin color channel $s$ (median color of the face) and the incomplete model image $y_i$ and generates the final image $y'$ of the model wearing garment $x$. Two modules are trained jointly.
   }
\vspace{-0.5cm}  

\label{fig:meta_figure}
\end{figure*}

\section{Outfit Visualization Net}
\label{sec:network}
We propose Outfit Visualization Net (OVNet) to generate images of a model (person) wearing multiple garments (outfit), faithfully capturing the garments details and the interactions between them. OVNet consists of two components trained separately: a Semantic Layout Generator $G_{layout}$ and a Multi-Warps Garment Generator $G_{garment}$. 

{\bf Semantic Layout Generator $G_{layout}$} predicts semantic layout $m'$ (in the form of segmentation map) conditioned on a garment image $x$, a pose map $p$ of the model and an incomplete layout $m_i$ (more details in appendix). This incomplete layout $m_i$ hides relevant semantic layout classes. For example, when generating the incomplete layout $m_i$ for a top, we take the ground truth layout $m$ and set the top, neckline, and arm classes to the background class. The generated layout $m'$ is then used to guide the image generation.

{\bf Multi-Warps Garment Generator $G_{garment}$} takes in the garment image $x$, the predicted full layout $m'$ and the model image $y$, and produces an image $y'$ with model $y$ wearing garment $x$. $G_{garment}$ only modifies one garment on the model at a time. Thus, garments of other categories remain unchanged from $y$ to $y'$.

Using our formulation, synthesizing an outfit requires multiple sequential operations, with each operation swapping a single garment. Compared to Neuberger \etal's~\cite{Neuberger_2020_CVPR} formulation, which is forced to generate a complete layout per inference, our formulation enables users to modify a single garment at a time, leaving the rest of the image untouched (Figure~\ref{fig:outfit-visualization}). Having this property benefits the user experience, as most people modify an outfit one piece at a time. The proposed method can be adopted to all application scenarios {\bf (a), (b), and (c)} (from the Intro~\ref{sec:intro}).

\subsection{The Semantic Layout Generator}\label{sec:layout_generator}

When synthesizing a person image, it is common practice to produce a semantic layout as structural constraints can guide the image generation \cite{Dong2018SoftGatedWF,Lassner:GeneratingPeople:2017,hsiao2019fashionplus,zhu2017be,Han_2019_ICCV,tprvton} and we follow a similar procedure. 
To train the layout generator, we obtain pairs of garment images $x$ and model images $y$ wearing $x$. From $y$, we obtain the semantic layout $m$ using off-the-shelf human parsing models~\cite{li2019self} and the pose map $p$ using OpenPose~\cite{wei2016cpm, 8765346, simon2017hand, cao2017realtime} (Figure~\ref{fig:meta_figure} top left). Based on the garment category of $x$, we produce an incomplete layout $m_i$ by setting the garment prediction classes as the background class. A full list of semantic categories and the detailed procedure for producing the incomplete layout $m_i$ for different categories of garments are in Appendix.

The layout generator takes in the incomplete layout concatenated with the pose and the garment as input, and learns to predict the original layout $m'= G_{layout}([x, m_i, p])$. Because skip connections propagate information from the input to the output, we use a U-Net architecture to retain information from $m_i$ in the output $m'$. The network is trained using a pixel-wise cross-entropy loss and a LSGAN~\cite{arxiv1611.04076} loss to encourage the generated semantic layouts to resemble real semantic layouts. The total training loss for $G_{layout}$ can be written as

\vspace{-0.5cm} 
\begin{equation}
\mathcal{L}_{layout} = \lambda_{1}\mathcal{L}_{CE} + \lambda_{2}\mathcal{L}_{GAN}
\end{equation}
\vspace{-0.5cm}

where $\lambda_{1}$ and $\lambda_{2}$ are the weights for each loss. Because the argmax function is non-differentiable, we adopt the Gumbel softmax trick \etal~\cite{45822} to discretize the layout generator's output such that the gradient generated by the discriminator can flow back to the generator. 

\begin{figure}
\begin{center}
   \includegraphics[width=0.99\linewidth]{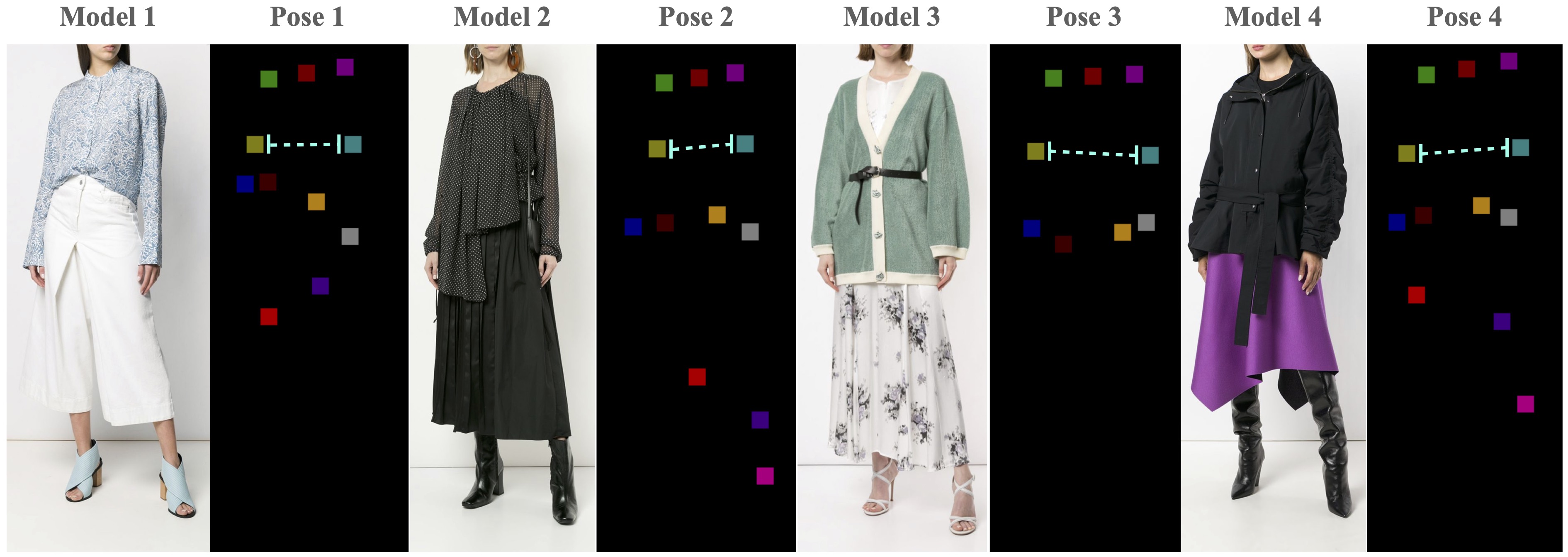}
\end{center}
           \vspace{-0.3cm}  
    \caption{We notice that the human pose annotation from OpenPose embeds information differently depending on the type of garment. For example, the pose predictor consistently predicts wider distance between shoulder and elbow anchor for models wearing coats (3, 4) than models wearing shirts (1, 2), despite both models having similar posture and body shape. This implies that the pairing between pose and garments can influence the predicted layout.
    }  
    \label{fig:model_pose}
       \vspace{-0.5cm}  

\end{figure}

During experiments, we observed that the type of garment a model is wearing greatly influences pose prediction results, as in Figure~\ref{fig:model_pose}. For example, between models with highly similar poses, one wearing a jacket and another one wearing a t-shirt, we observe vastly different pose predictions. Also, because we train the network to reconstruct the ground truth semantic layout conditioned on garment and pose, the pose representation may impose a prior on the type of garment to expect. This sometimes leads to errors during inference. As in Figure~\ref{fig:good-bad_matching}, when there is a mismatch between the provided garment (a tank) and what the pose representation implies (a jacket), the layout generator may output a layout that doesn't respect the garment shape. In Section~\ref{sec:garment_pose_matching}, we propose a garment-pose matching procedure to alleviate this issue.

\subsection{Multi-Warps Garment Generator}\label{sec:multi_garment_generator}


Our garment generation network $G_{garment}$ (Figure~\ref{fig:meta_figure} right) takes in a garment image $x^c$ of class $c$ (write as $x$ for simplicity), a model image $y$ and a segmentation mask $m^c$ covering the target garment's class $c$ on the model image $y$, and generates a realistic synthesized image of the model wearing the provided garment. $G_{garment}$ consists of two modules: (a) a warper to create {\bf $k$ specialized warps}, by aligning the garment image $x$ with the semantic layout $m^c$ of the garment class; (b) an inpainting module to generate the final image leveraging the warps, the semantic layout $m$, the skin color of the model $s$ (median color of the face), and the incomplete model image $y_i$ where the target garment, skin, and background are masked out. Unlike prior works~\cite{han2017viton,wang2018toward,Han_2019_ICCV,tprvton} that learn a single warp with high degrees of freedom to align garments, our method learns a family of warps, each specializing on certain features. The inpainting network is fed all the warps and learns to combine them by choosing features to look for from each warp, as it is trained jointly with the warper. 


The {\bf Warping Module } resembles a spatial transformer network~\cite{NIPS2015_5854}. First, a regressor takes in the garment image $x^c$ and the mask $m^c$ as input, and regress $k$ sets of spatial transformation parameters ${\theta_1...\theta_k}$. Then, it generates a grid for each of the transformation parameters, and samples grids from the garment image $x$ to obtain $k$ warps $w_1..w_k$ where $ w_1=\mathcal{W}(x,\theta_1)$. The warps are optimized to match the garment worn by the target model $m^c \otimes y$ using per pixel $\mathcal{L}_1$ loss. Inspired by ~\cite{Han_2019_ICCV}, we impose a structure loss to encourage the garment region $z$ (a binary mask separating garment and background as in Figure~\ref{fig:pose_germent_matching}) of $x$ to overlap with the garment layout of the garment mask $m^c$ on the model after warping. The warping loss can be written as:
\begin{equation}
\mathcal{L}_{warp}(k) = \\|\mathcal{W}(x,\theta) - (m^c \otimes y)\\| + \beta \\|\mathcal{W}(z,{\theta}_k) - m^c \\|
\end{equation}


where $\beta$ controls the strength of the structure loss. This loss is sufficient to train a single warp baseline method. The choice of warper here is unimportant, and in our implementation, we use affine transformation with $6$ parameters.

{\bf Cascade Loss:} With multiple warps, each warp $w_j$ is trained to address the mistakes made by previous warps $w_i$ where $i < j$. For the $k$ th warp, we compute the minimum loss among all the previous warps at every pixel location, written as
\begin{equation}
\mathcal{L}_{warp}(k) = \frac{\sum_{u=1, v=1}^{W,H}{ min(\mathcal{L}_{warp}(1)_{(u, v)} .. \mathcal{L}_{warp}(k)_{(u, v)})}}{wh}
\end{equation} where $u,v$ are pixel locations; $W,H$ are the image width and height; and $\mathcal{L}_{warp}(k)_{(u, v)}$ is the loss of the $k$th warp at pixel location $u,v$. The cascade loss computes the average loss across all warps. An additional regularization term is added to encourage the transformation parameters of all later warps to stay close to the first warp.
\begin{equation}
\mathcal{L}_{casc}(k) = \frac{\sum_{i=1}^k{\mathcal{L}_{warp}(i)}}{k} + \alpha\frac{\sum_{i=2}^k{\|\theta_k - \theta_1\|^2}}{k-1}
\end{equation}
The cascade loss enforce a hierarchy among all warps, making it more costly for an earlier warp to make a mistake than for a later warp. This prevents oscillation during the training (multiple warps competing for the same objective). 

The idea is comparable with boosting -- using multiple simple warpers (weak learners), each with a small degree of freedom but can handle complex geometric shape when combined. Warpers interact with each others differently compared to classifiers. Concatenating multiple warps channel-wise allows a generator to reason about the geometrics while also leveraging the parallelism of the computation (less latency). Training end-to-end allows all warps to share gradients, making it possible for warps to adjust according to each other and the image generator to guide the warpers.

The {\bf Inpainting Module} concatenates all the warps $w_1..w_k$, the semantic layout $m$ (or $m'$ during inference), and the incomplete image $y_i$ as input, and outputs the final image $y'$ of model $y$ wearing garment $x$. 
This is different from a standard inpainting task because the exact content to inpaint is provided through the input channels. We use a U-Net architecture to encourage copying information from the input. The network is trained to reconstruct the ground truth image using a per pixel $\mathcal{L}_{1}$ loss, a perceptual loss~\cite{Johnson2016Perceptual}, and a Spectral Norm GAN with hinge loss~\cite{miyato2018spectral}. The total loss for training $G_{garment}$ with $k$ warps is written as 
\begin{equation}
\mathcal{L}_{garm}(k) = \gamma_{1}\mathcal{L}_{casc
}(k) + \gamma_{2}\mathcal{L}_{1} + \gamma_{3}\mathcal{L}_{perc} + \gamma_{4}\mathcal{L}_{GAN}
\end{equation}
where $\gamma_{1}$, $\gamma_{2}$, $\gamma_{3}$ and $\gamma_{4}$ are the weights for each loss.


\section{Garment-Pose Matching}\label{sec:garment_pose_matching}

\begin{figure}
\begin{center}
   \includegraphics[width=0.9\linewidth]{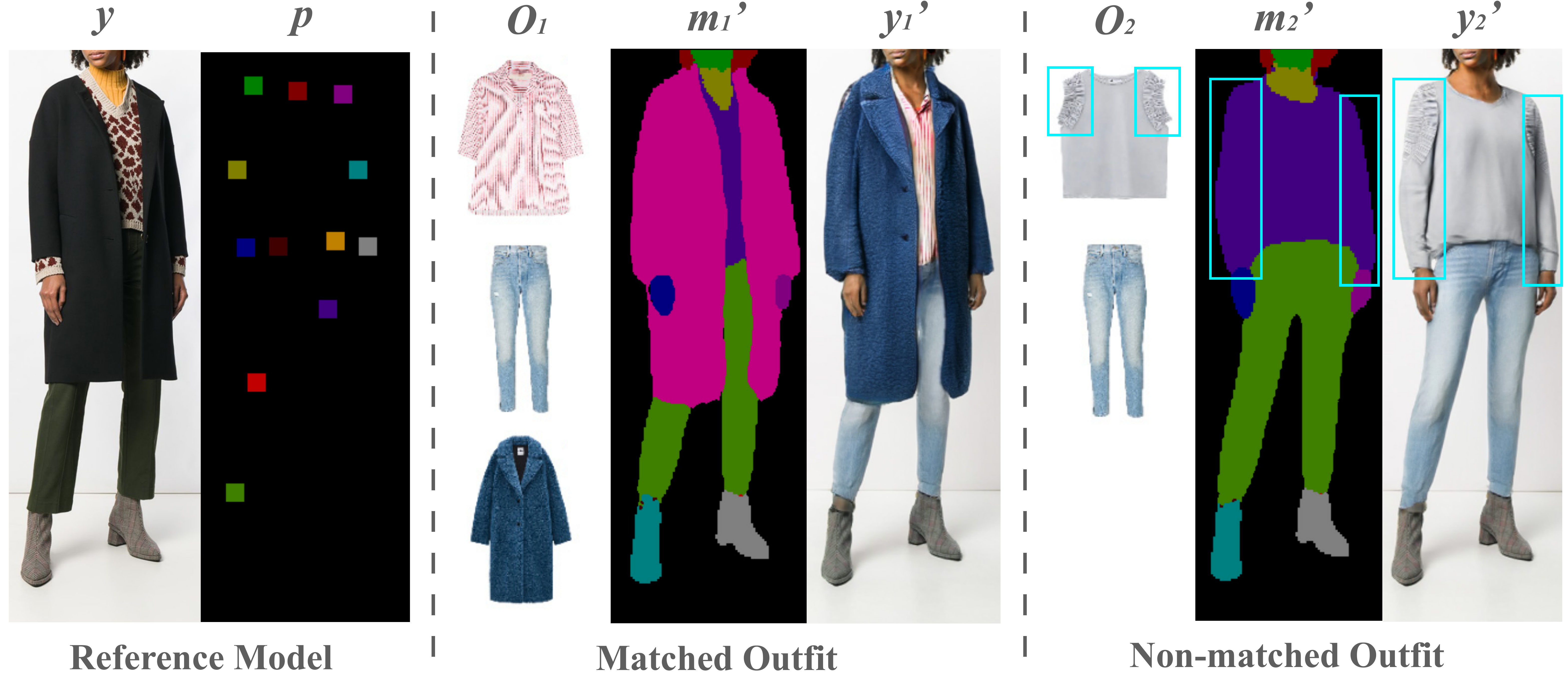}
\end{center}
           \vspace{-0.3cm}  
    \caption{This figure shows an example result from a matched garment-pose pair versus a non-matched pair. A model $y$ with extracted pose $p$ is fed two different outfits $O_1$ and $O_2$. The garments in $O_1$ match with the shape of garments worn by the original model $y$, thus results in an accurate layout prediction $m_1'$ and output $y_1'$. In contrast, the sleeveless tank in $O_2$ does not match with pose $p$, thus was wrongly generated with sleeves in $y_2'$.}    
    \label{fig:good-bad_matching}
           \vspace{-0.3cm}  
\end{figure}

    While our Outfit Visualization Network and other prior works~\cite{Han_2019_ICCV, tprvton} support shape changes (e.g., skirt to pants, long sleeve to short sleeve), we notice that semantic layout generators strongly favor certain garment-model(person) pair over others. The root cause is because the pose detection results are heavily biased by garments (Figure~\ref{fig:model_pose}). For example, the pose representation extracted from a person wearing a long dress has attributes (e.g., odd position of the feet, wide legs, etc.) that hint to the generator to expect a long dress, Figure~\ref{fig:model_pose}. Thus, during inference, putting a different garment (e.g. trousers) on this model will cause problems (Figure~\ref{fig:good-bad_matching}), because the garment and pose are always extracted from the same person during training. Fully addressing this problem may require improving pose representations and is left as a future direction.

To overcome such deficiency, we propose that choosing a suitable model for a given set of garments will result in better quality generation compared to using a random model. The strategy can be adopted in application scenarios {\bf (a) and (b)} (from the Intro~\ref{sec:intro}) where we are not forced to operate on a fixed model image. The general relationship between pose and garment is hard to capture, but we expect a garment $x_a$ to work well with its paired model $y_a$. Also, because shape is the only relevant attribute to the semantic layout, we expect a garment $x_b$ with similar shape as $x_a$ to work better with $y_a$ than a garment $x_c$ with a different shape. We want an embedding to capture such property. 

\begin{figure}
\begin{center}
   \includegraphics[width=.95\linewidth]{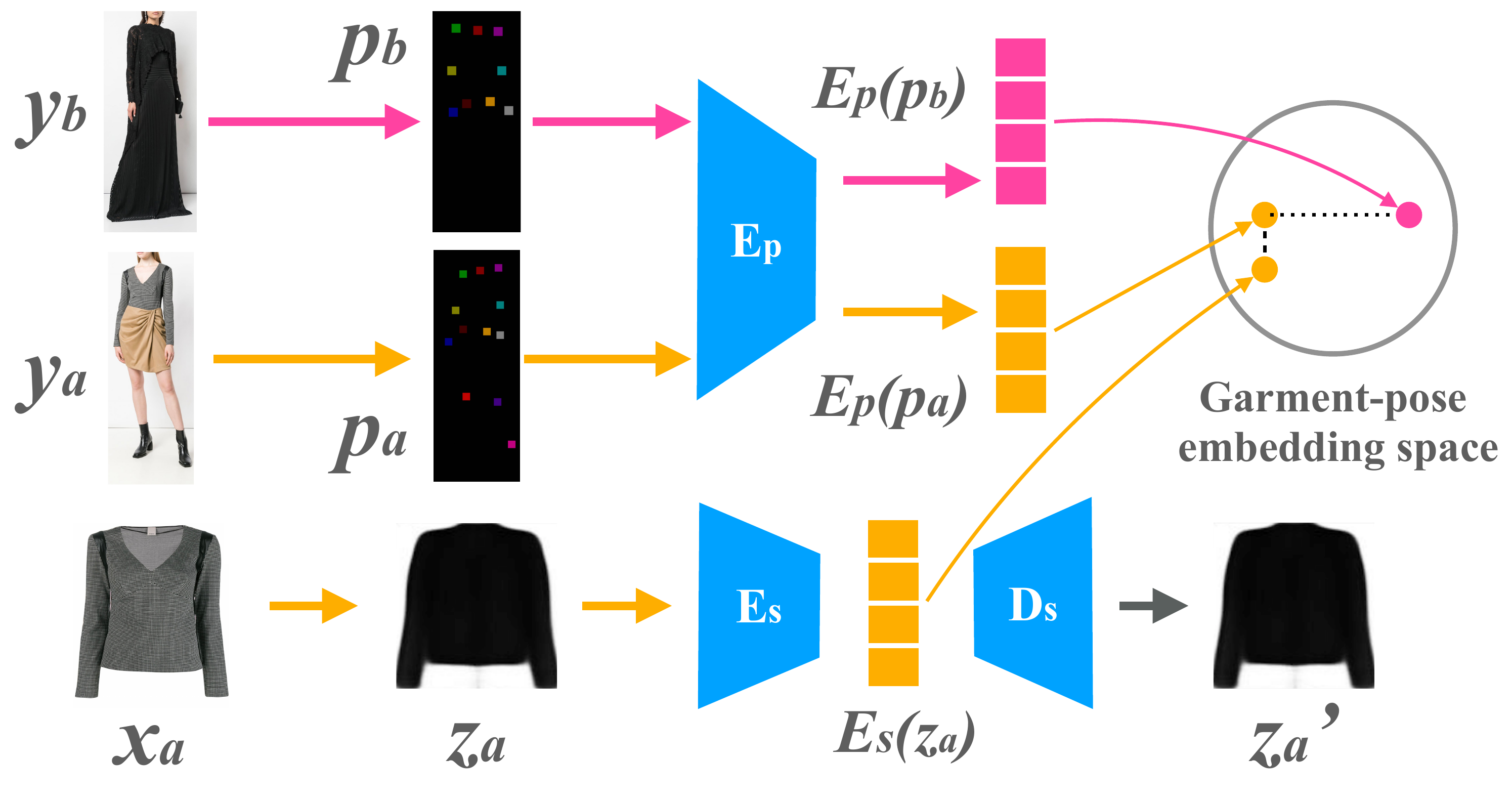}
\end{center}
\vspace{-0.5cm}

   \caption{ This figure shows the training procedure for the garment-pose matching embedding. We obtained a foreground mask $z_a$ from garment image $x_a$, and learn a shape Auto-encoder $\{E_s, D_s\}$ to produce a shape embedding. The pose $p_a$ from the corresponding model $y_a$ is embedded closer to $z_a$ than a random pose $p_b$. This only works for scenarios {\bf (a) and (b)} (from Intro~\ref{sec:intro})
   }
   \vspace{-0.5cm}
\label{fig:pose_germent_matching}
\end{figure}

To train the garment-pose embedding, we first learn a Garment Shape Auto-encoder $\{E_s, D_s\}$ to obtain a condensed garment shape representation (Figure~\ref{fig:pose_germent_matching}). We use the garment's foreground mask $z$ (a binary mask of 1's for foreground and 0's for background) as input, and train the Auto-encoder to reconstruct the garment mask $z' = D_s(E_s(z))$ using mean squared error as the reconstruction loss. Additionally, we apply $\mathcal{L}_2$ normalization on the Auto-encoder's embedding space and we regard the data encoding $E_s(z)$ as an embedding for garment shape. Subsequently, we learn a pose encoder $E_p$ to project Openpose map $p$ into the shape embedding space. $E_p$ is trained using a Triplet loss $\mathcal{L}_{triplet}$~\cite{conf/cvpr/SchroffKP15} to encourage $p_a$ and $z_a$ with an identical garment $a$ to have a closer embedding to each other compared to a randomly sampled pose $p_b$ by a margin of $\alpha$. The full training loss is written as
\begin{multline}
\mathcal{L}_{match} = \|D_s(E_s(z_a)) - z_a\|^2 + \\ \mathcal{L}_{triplet}(E_s(z_a), E_p(p_a), E_p(p_b))
\end{multline} Because the same pose may correspond to garments of multiple categories, we train a set of specific pose encoders $\{E_p^{c_1}..E_p^{c_n}\}$ for each garment category $c \in C$.

At inference time, we search for a set of suitable poses given a query outfit $O = \{z^{c_1}, ..., z^{c_m}\}$ (a set of garments of different categories). We compute the distance between the outfit $O$ and a pose $p$ as the maximum distance between the shape embedding of any garment in the outfit and the pose embedding: $d(O, p) = max(\{ \|E_s(z^{c_i}) - E_p^{c_i}(p)\|^2, z^{c_i} \in O\})$. The images whose pose have the shortest distances to the query outfit are preferably chosen.

\section{Experiments}\label{sec:experiments}

\subsection{Datasets \& Experiment Setup}
\label{sec:setup}
Because publicly available try-on datasets do not include rich garment categories, we experiment on a new dataset of 321k fashion products scraped from e-commerce websites, containing all the available garment categories. Each product includes a neutral garment image (front-view, laying flat, plain background), and a model image (single person, front-view). Garments are grouped into four types (top, bottoms, outerwear, or full-body). We randomly split the data into 80\% for training, 5\% for validation and 15\% for testing. Because the model images do not come with body parsing annotation, we use off-the-shelf human parsing models~\cite{li2019self} to generate semantic layouts as training labels.

We also compare with prior work on the established VITON dataset~\cite{han2017viton}. Note we do not compare with single-garment try-on methods on the new multi-category dataset because single-garment try-on methods do not work reasonably on our dataset, we expand on this in our supplementary. Because the original VITON test set consists of only 2,032 garment-model pairs (insignificant for computing FID), we resample a larger test set of 50k mismatched garment-model pairs, following the procedure of the original work~\cite{han2017viton}. To quantify the effect of garment-poses on generation quality, we create another resampled test set where garment-model pairs are selected using our Garment-Pose matching procedure: every garment in the original test set is paired with its 25 nearest neighbor models in the pose embedding space. 

Other details about network architectures, training procedures and hyper parameters are provided in the Appendix.

 \begin{figure}
\begin{center}
   \includegraphics[width=.95\linewidth]{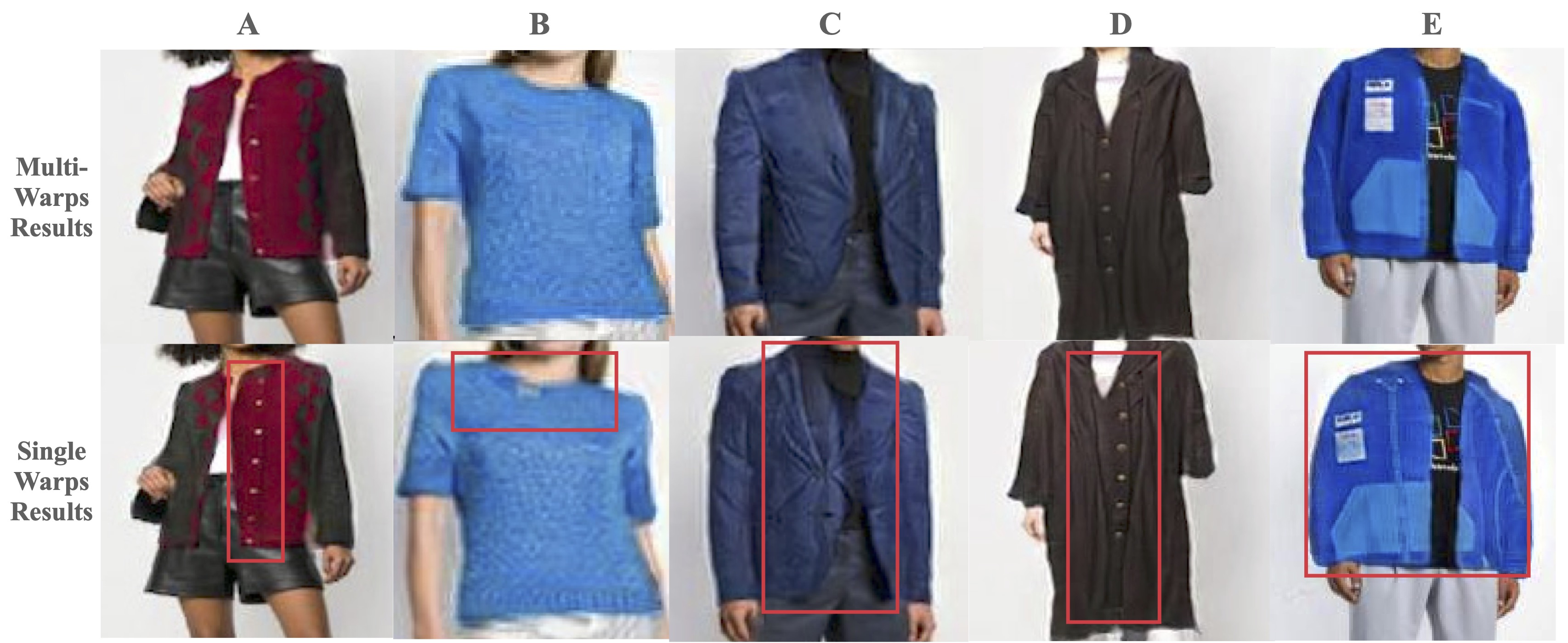}
\end{center}
\vspace{-0.3cm}  
   \caption{ The figures shows qualitative comparison between using multiple (2) warpers and a single warper. Note for single warp: the buttons are in the wrong place in A and D; problems with sleeve boundaries in E; a severe misalignment in C; a misplaced tag in B. All problems are fixed in multi-warp results.
}
   \label{fig:multi-warp-qualitative}
          \vspace{-0.3cm}  

\end{figure}

\begin{table}[]
\begin{center}
\begin{tabular}{c|c|c}
\hline
Methods           & SSIM & IS \\ \hline
VITON~\cite{han2017viton}            & .783  & 2.650        \\ 
CP-VTON~\cite{wang2018toward}            & .745  & 2.757         \\ 
GarmentGAN~\cite{Raffiee2020GarmentGANPA} & -  & 2.774 	  \\
VTNFP~\cite{YuVTNFPAI}            & .803  & 2.784         \\ 
SieveNet~\cite{Jandial2020SieveNetAU}          & .766  & 2.820         \\ 
ClothFlow~\cite{Han_2019_ICCV}            & .841  & -         \\ 
ACGPN~\cite{tprvton}            & .845  & 2.829         \\ 
Ours (4 warps)         & {\bf .852}  & {\bf 2.846 }         \\ \hline
\end{tabular}
\end{center}
\vspace{-0.2cm}
\caption{This table compares SSIM~\cite{1284395} and IS~\cite{NIPS2016_8a3363ab} (larger is better) reported on the original VITON test set. Results show that our garment generation pipeline outperforms prior works.
\vspace{-0.1cm}}  
\label{table:is_ssim}
\end{table}

\begin{table}[]
\begin{center}
\begin{tabular}{c|c|c}

\hline
Methods           & Random Pairs & Matched Pairs \\ \hline
CP-VTON~\cite{wang2018toward}            & 15.11  & 13.42         \\ 
ACGPN~\cite{tprvton}           & 11.13  & 9.03          \\ \hline
Ours (4 warps)          & 9.81   & \textbf{7.02} \\ \hline
\end{tabular}
\end{center}
\vspace{-0.2cm}
\caption{This table compares the FID$_\infty$~\cite{chong2019effectively} score on two resampled test sets (see Sec.~\ref{sec:setup}), one randomly sampled and the other using our pose-garment matching. Results show that choosing compatible pairs yield significantly improves to all try-on methods.
\vspace{-0.6cm}}
\label{table:fid} 
\end{table}

\subsection{Quantitative Results}
 \begin{figure}
\begin{center}
\vspace{-0.5cm}  
	\includegraphics[width=0.99\linewidth]{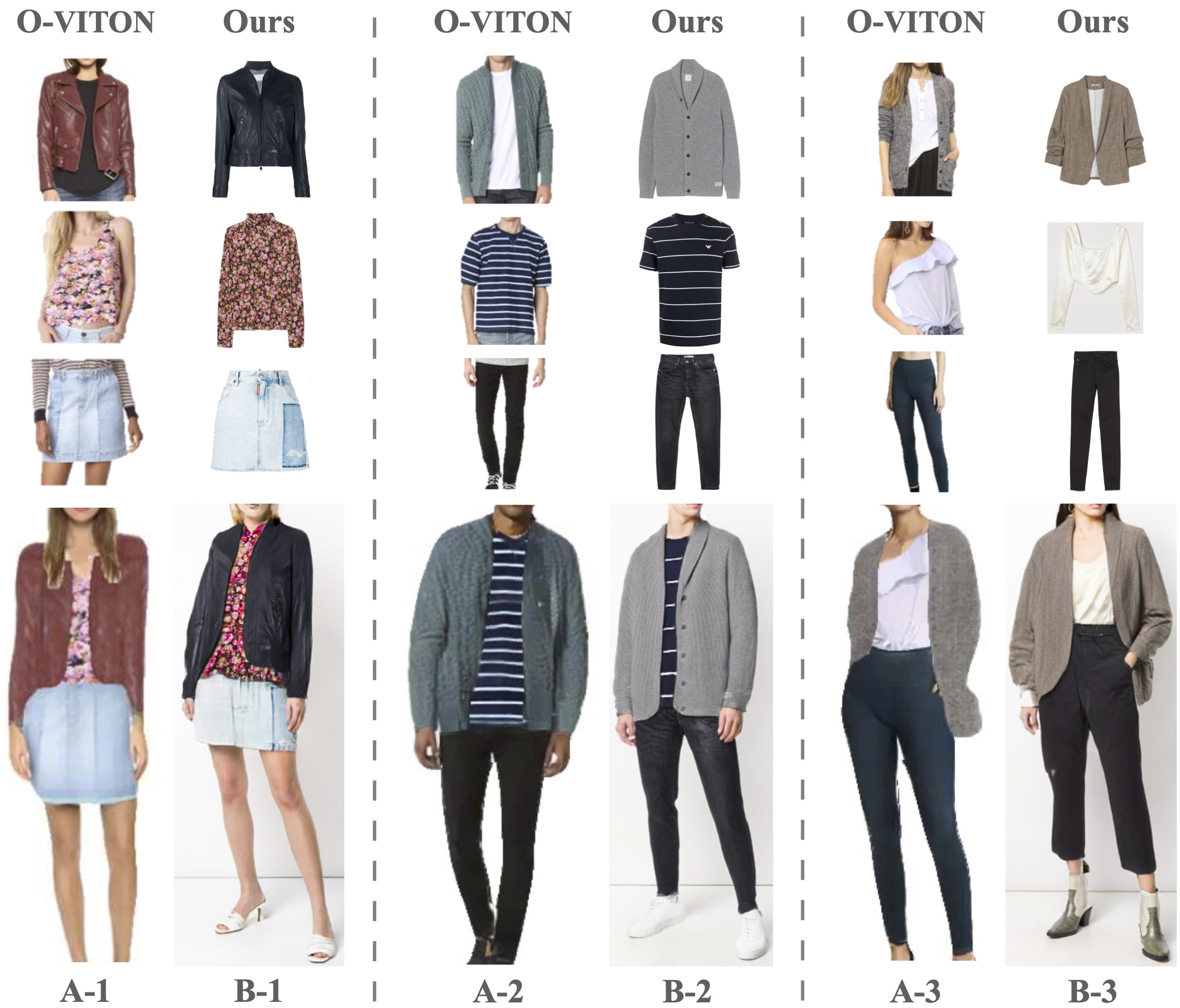}
    \caption{We compare visual results between O-VITON~\cite{Neuberger_2020_CVPR} and ours. The top rows show the garments in the oufit and the bottom row shows the generated try-on results. For fair comparison, we found garment images that most closely resemble the garments chosen in~\cite{Neuberger_2020_CVPR} in terms of style, color, and texture. Image results for O-VITON are directly taken from their paper. There are substantial difference in quality between results. The unnaturally flat torso and uneven shoulders of A-1 are not present in B-1. In A-2, the buttons on the jacket are distorted/missing, whereas B-2 represents them accurately. In A-3, the jacket and top lack realism due to missing creases, folds, and bumps compared to B-3. Properties of the arms are also kept intact in B-3. (See Appendix for more)}    
    \label{fig:amazon_compare}
    \end{center}
\vspace{-0.5cm}  
\end{figure}

Following prior works, we report SSIM~\cite{1284395} and IS~\cite{NIPS2016_8a3363ab} scores on the original VITON test set~\cite{han2017viton}. As shown in Table~\ref{table:is_ssim}, our multi-warp generation pipeline outperforms prior works in both metrics. Additionally, while Frechet Inception Distance (FID)~\cite{heusel2017gans} is commonly used to evaluate generated image quality~\cite{brock2018large,zhang2018self,karras2019style}, Chong~\etal ~\cite{chong2019effectively} recently showed that FID is biased and proposed an extrapolation to an unbiased score (FID$_\infty$). We adopt FID$_\infty$ in our work over FID. Results from WUTON~\cite{Issenhuth2020DoNM} are excluded because their experiments were conducted on a different dataset. 

\begin{table}[]
\begin{center}
\setlength\tabcolsep{5pt} 
\begin{tabular}{c|cccc|c}
\hline
warp          & bottoms & full-body & tops & outerwear & overall \\ \hline
1            & 1.930  & 4.461 & 2.489 & 2.233 & 1.577        \\ 
2            & 1.472  & 2.207 & 1.215 & 1.349 & .927       \\ 
4           & 1.461  & 2.069 & \textbf{1.163} & 1.328 & .874  \\ 
8          & \textbf{1.458}   & \textbf{2.057} & 1.165	 & \textbf{1.323} & \textbf{.872} 	\\ \hline
\end{tabular}
\vspace{-0.2cm}
\end{center}
\caption{This table reports the FID$_\infty$~\cite{chong2019effectively} score (smaller is better) of our method on the new multi-category dataset. We compare the performance between using different numbers of warps. Results shows that using more warps significantly increase performance.
\vspace{-0.5cm}} 
\label{table:multi_cat}
\end{table}

Neuberger \etal's~\cite{Neuberger_2020_CVPR} is the only known prior work that supports multi-garment try-on. However, quantitative comparison is impossible as (1) their code and dataset are not released and (2) their formulation uses images of people wearing garment rather than neutral garment images. Instead, we compare with them qualitatively (Figure~\ref{fig:amazon_compare}). 

To evaluate our garment-pose matching procedure, we run OVNet and prior methods with released implementations~\cite{wang2018toward, tprvton} on two resampled test sets of 50k pairs, one sampled using the garment-pose matching procedure and the other sampled randomly. 
We report results in Table~\ref{table:fid}. Using garment-pose matching significantly improves results for all methods, even those that are designed to accept arbitrary garment-model pairs (ours and ACGPN~\cite{tprvton}). Additionally, our garment generation pipeline shows consistently better FID$_\infty$ scores compared to other methods.

Table~\ref{table:multi_cat} reports the FID$_\infty$ for our method on the multi-category test dataset using different number of warps. Using more warps substantially improves the performance on all garment categories, with diminishing returns as it increases. We set the number of warps to 4.

\subsection{Qualitative Comparison}

We show comprehensive qualitative examples of our method. In Figure~\ref{fig:multi-warp-qualitative}, we show how multiple warpers can significantly improve and correct the details. In Figure~\ref{fig:header}, we show examples of how garment details are realistically captured: patterns (A-1), shadows (B-1, C-2, D-2), hemlines (C-2), buttons (B-1) and numerous other features are all accurately represented (refer to figure for more details).

In Figure~\ref{fig:multiple_models}, we show that our method can generate the same outfit selection on a diverse set of models and poses (e.g. different stances, skin colors, and hand/arm positions). The garments' properties are consistent across all models, suggesting that the network has learned a robust garment representation. Pay attention to Pose 2 \& 5 when the hands are in the pockets; the jacket/jean pocket plumps up and the sleeve interacts seamlessly with the pocket. These realistic details are likely results of using a GAN loss.

Finally in Figure~\ref{fig:amazon_compare}, we compare our results to O-VITON~\cite{Neuberger_2020_CVPR}, the state-of-the-art in multi-garment try-on. Compared to O-VITON, our method applies clothes more naturally onto models (A-1 vs B-1), localizes buttons more accurately (A-2 vs B-2), and generates more realistic textures and more convincing fabric properties (A-3 vs B-3).

 \begin{figure}
\begin{center}
   \includegraphics[width=0.95\linewidth]{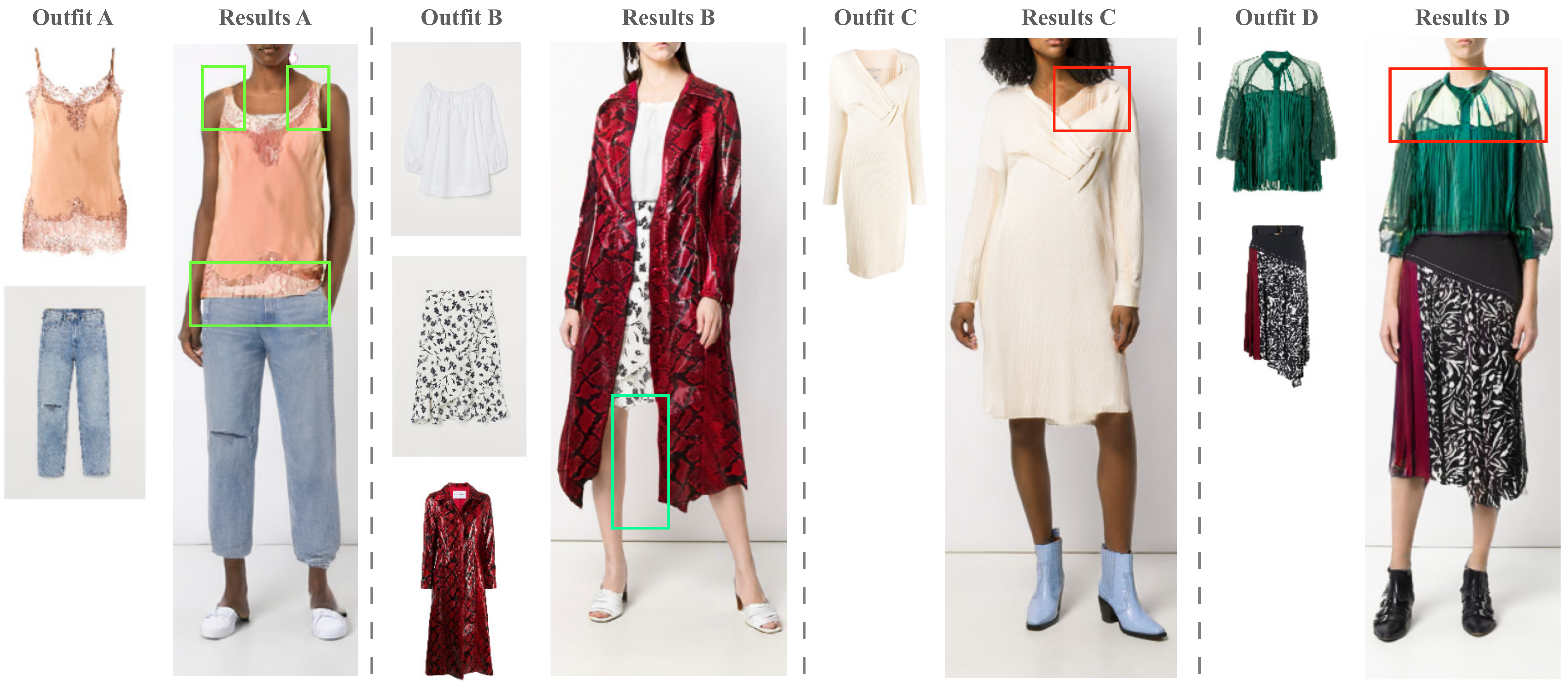}
\end{center}
\vspace{-0.3cm}  

    \caption{Our method has forgiving failure modes. When it fails, it still outputs an image of the person wearing realistic garments, but with misrepresented attributes. In A, it turns spaghetti straps into thick straps, and has difficulty with laces; in B, the coat is generated as open-back; the asymmetrical neckline in C is turned into panels; and transparency is not captured in D.}    
    \label{fig:bad_examples}
           \vspace{-0.5cm}  

\end{figure}

We also show common mistakes made by our method in Figure~\ref{fig:bad_examples}. Our mistakes tend to be quite forgiving, resulting in inaccurate but realistic interpretations of the outfits. These failures are caused by inaccurate layout predictions.

\begin{figure}
\begin{center}
   \includegraphics[width=0.95\linewidth]{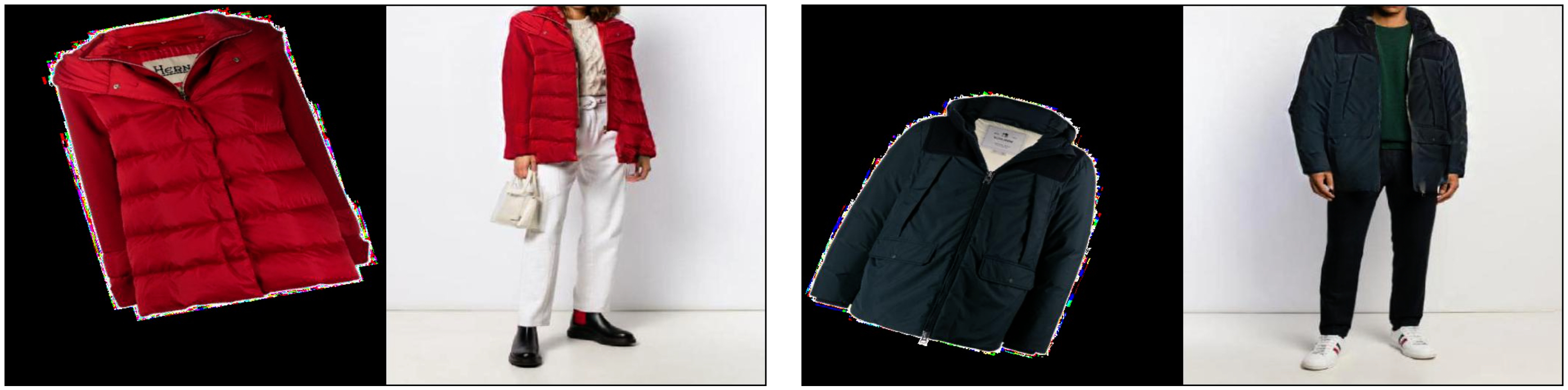}
\end{center}
\vspace{-0.3cm}  

   \caption{ Two synthesized images that 70\% of the participants in the user study thought were real. 
Note, e.g., the shading, the wrinkles, even the zip and the collar.
   }
\label{fig:user-study-examples}
\vspace{-0.3cm}  

\end{figure}

\begin{table}
\begin{tabular}{l|l|lll}
\hline
 &Participants & Acc & FP & FN \\
\hline
Crowds & 31 & 0.573 & { \bf 0.516}& 0.284 \\
Researchers & 19 &0.655 & { \bf 0.615} & 0.175\\
\hline
\end{tabular}
\vspace{0.1cm}
\caption{The user study results show that participants have difficulty distinguishing between real and synthesized images. 51.6\% and 61.5\% of fake images were thought to be real by crowds and researchers, respectively. Some of the real images were marked as fake, suggesting participants were actively trying to spot flaws.
\vspace{-0.5cm}}
\label{table:user_study}
\end{table}

To further substantiate the quality of our image generation from a provided layout, we perform a user study to verify how often users can distinguish synthesized images from real images. A user is presented with an image of the product and an image of a model wearing the product. The user is then asked if the image of the model wearing the product is real or synthesized. 

The results of our case study show that users are mostly fooled by our images; there is a very high false-positive rate (i.e. synthesized image is marked real by a user; Table~\ref{table:user_study}). Figure~\ref{fig:user-study-examples} shows two examples of synthesized images that 70\% of participants reported as real. These are hard outerwear examples with multiple segmented regions and complicated shading. Nevertheless, our method manages to generate high quality synthesized images that consistently fool users. See supplementary material for the complete settings and results of the user study.

\section{Conclusion \& Discussions}
In this work, we propose a systematic method to enable outfit-level generation with realistic garment details. Several design choices are crucial. (1) We operate on neutral garment images rather than images of garments worn by models. We believe using neutral product images is more accessible for consumers and readily provided by clothing brands, making our solution easily adoptable. (2) Using warping is important toward accurately preserving geometric textures. Warping multiple garments with complicated shapes is extremely challenging, and we are the first to demonstrate success in generation of all garment categories through warping. (3) Even though, our try-on generation pipeline (as well as others) support arbitrary pairs of garment and model images, we demonstrate that it is highly advantageous to carefully choose the pair when possible.

Despite the success, our method can be improved in many aspects. Our method can handle variations in body pose and skin tone, but not body shape. Enabling body shape variations would get us one step closer to achieving the difficult goal of dressing garments directly on consumers' photos. For such a task, the main challenge lies in handling out of distribution user-uploaded photos. Additionally, enabling try-on for shoes, bags, and other accessories would make the outfit generation complete. 

{\small
\bibliographystyle{ieee_fullname}
\bibliography{egbib}
}

\newpage
\section{User Study}

The user study evaluates both the realism and accuracy of the generated image conditioned on correct layout. Because we want only aim to evaluate the image generator, we use the layout produced by the pre-trained human parser on the ground truth image (rather than the ones produced by the semantic layout generator). Because produced layout are often noisy, we choose examples where the layout is good, giving a fair representation of the top 20 percentile of our results. 

We ran our user study with two populations of participants (vision researchers tested with Form B, and randomly selected people tested with Form A). Images are displayed using the highest possible resolution (512x512) and each participant is primed with two real and fake examples before the study. Each participant is then tested with 50 examples (25 real and 25 fake), without repeating products. During the study, the garment image and the model image are shown side-by-side, giving subject an easy way to determine whether the synthesized image accurately represent the garment -- an important property for fashion e-commerce application.

For every model, we tested swapping a single garment. Each model in our dataset has a ground truth paired with only one garment (the other garments worn by the model are not matched with any product images). Both the real and fake images are shown with the same outfit. Form A and Form B mirror each other (i.e. if a garment is shown as the generated version in Form B, the real image will be shown in Form A).

The raw questions and responses are provided under the folder ``user study".


 \begin{figure}
\begin{center}
   \includegraphics[width=0.99\linewidth]{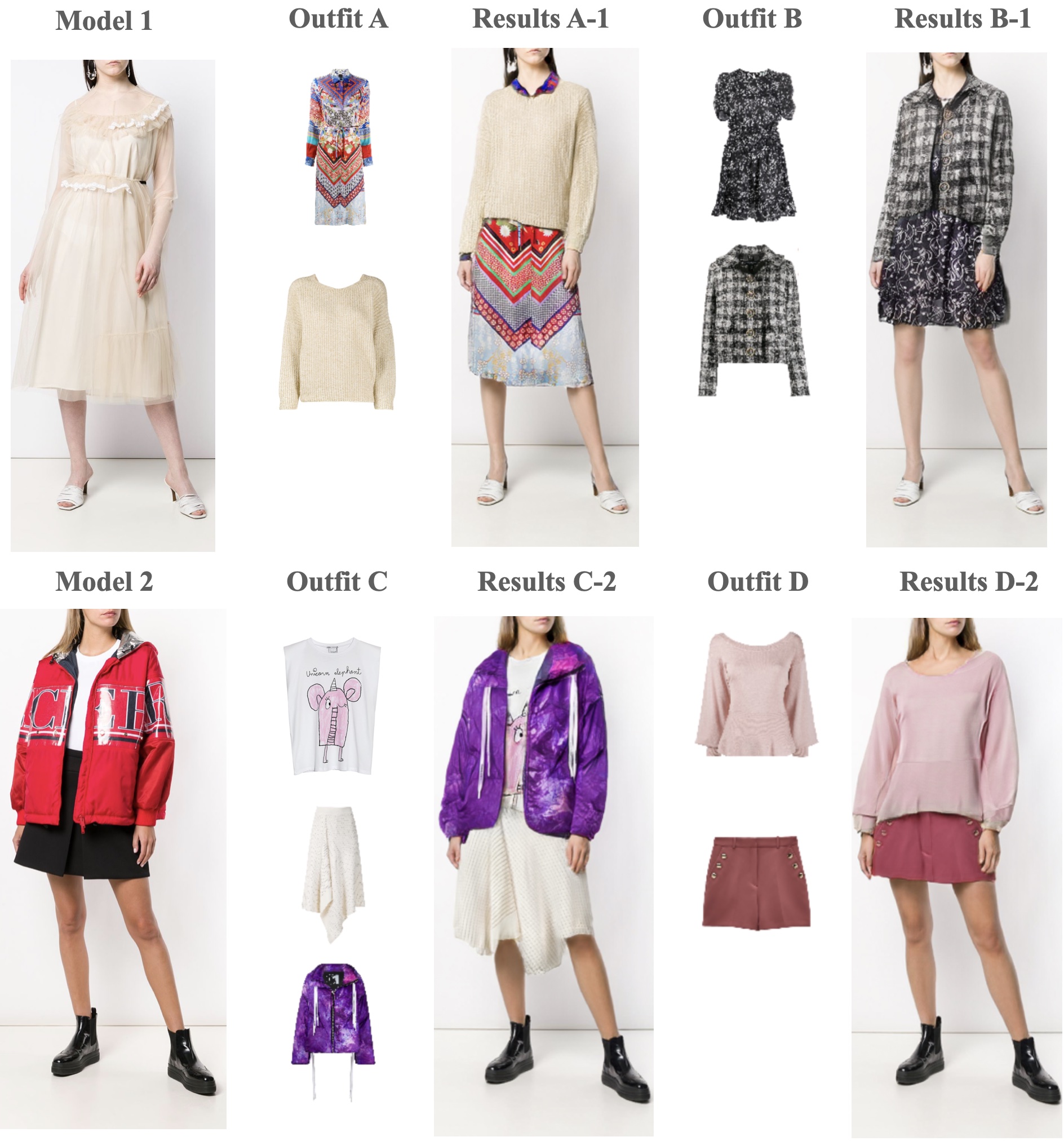}
\end{center}
           \vspace{-0.5cm}  
    \caption{A version of the Figure 1 without bounding boxes.}  
       \vspace{-0.5cm}  
          \label{fig:header}
\end{figure}

\section{Semantic Layout and Pose Representation}

\subsection{Semantic Layout}
We obtain our semantic layout using an off-the-shelf human parser~\cite{li2019self}. Our Semantic Layout has 15 semantic labels to represent different human parts and garments. These are background, hair, face, neckline, right arm, left arm, right shoe, left shoe, right leg, left leg, bottoms, full-body, tops, outerwear, and bags. Among these semantic labels, bottoms, full-body, tops, and outerwear are garment labels. The semantic segmentation mask is $m \in R^{H\times W \times15}$, where $H$ and $W$ correspond to the width and height of the image.

\subsection{Incomplete Layout}
During training, we create the incomplete layout $m_i$ by hiding the target garment labels and relevant skin labels (by setting these labels to the background class). For tops and outerwear, we hide tops, outerwear, left arm, right arm and neckline; for bottoms, we hide bottoms, left leg and right leg; for full-body, we hide full-body, left leg, right leg, left arm, right arm and neckline. All original channels are still outputted as the incomplete layout ($m_i \in R^{H\times W \times15}$).

\subsection{Pose Representation}
We first apply the pre-trained Openpose~\cite{8765346, wei2016cpm} model on the model images to obtain 18 key points. Following prior work~\cite{wang2018toward,Han_2019_ICCV,tprvton}, we convert the key points into a heatmap with 18 channels. Our key point heatmap becomes $p \in R^{H\times W \times18}$. In each channel, we draw a $5 \times 5$ square centered at each keypoint's coordinate and set all values in the square to one. If a key point is missing, the channel will be set to all zeros.

\section{Experiment Setups}

\subsection{Network Architecture}
For both the semantic layout generator $G_{layout}$ and the inpainting network of the multi-warp garment generator $G_{garment}$, we use a U-Net of 5 hidden layers. The channel sizes for the hidden layers are 64, 128, 256, 512, and 1,024 respectively. We downsample the image size by 2 at each layer, using bilinear interpolation. 

For the warper module of the multi-warp garment generator $G_{garment}$, we use a ResNet-18 model with ImageNet pre-trained weights as the backbone.

\subsection{Training Procedure}
We train our network using Adam Optimizer with a learning rate of $1e^{-4}$ for the semantic layout generator $G_{layout}$ and $2e^{-4}$ for the multi-warp garment generator $G_{garment}$. Both networks are trained on a Quadro RTX 6000 GPUs (24GB). $G_{layout}$ is trained for 50k steps with a batch size of 16. $G_{garment}$ is trained for 100k steps with a batch size of 8.

For training the $G_{layout}$, $\lambda_{1}$ and $\lambda_{2}$ are set to $1$ and $0.2$ respectively. For training the $G_{garment}$, $\gamma_{1}$, $\gamma_{2}$, $\gamma_{3}$ and $\gamma_{4}$ are set to $5$, $5$, $3$ and $1$, respectively.

\section{More Qualitative Comparisons}
\begin{figure*}
\begin{center}
   \includegraphics[width=0.88\linewidth]{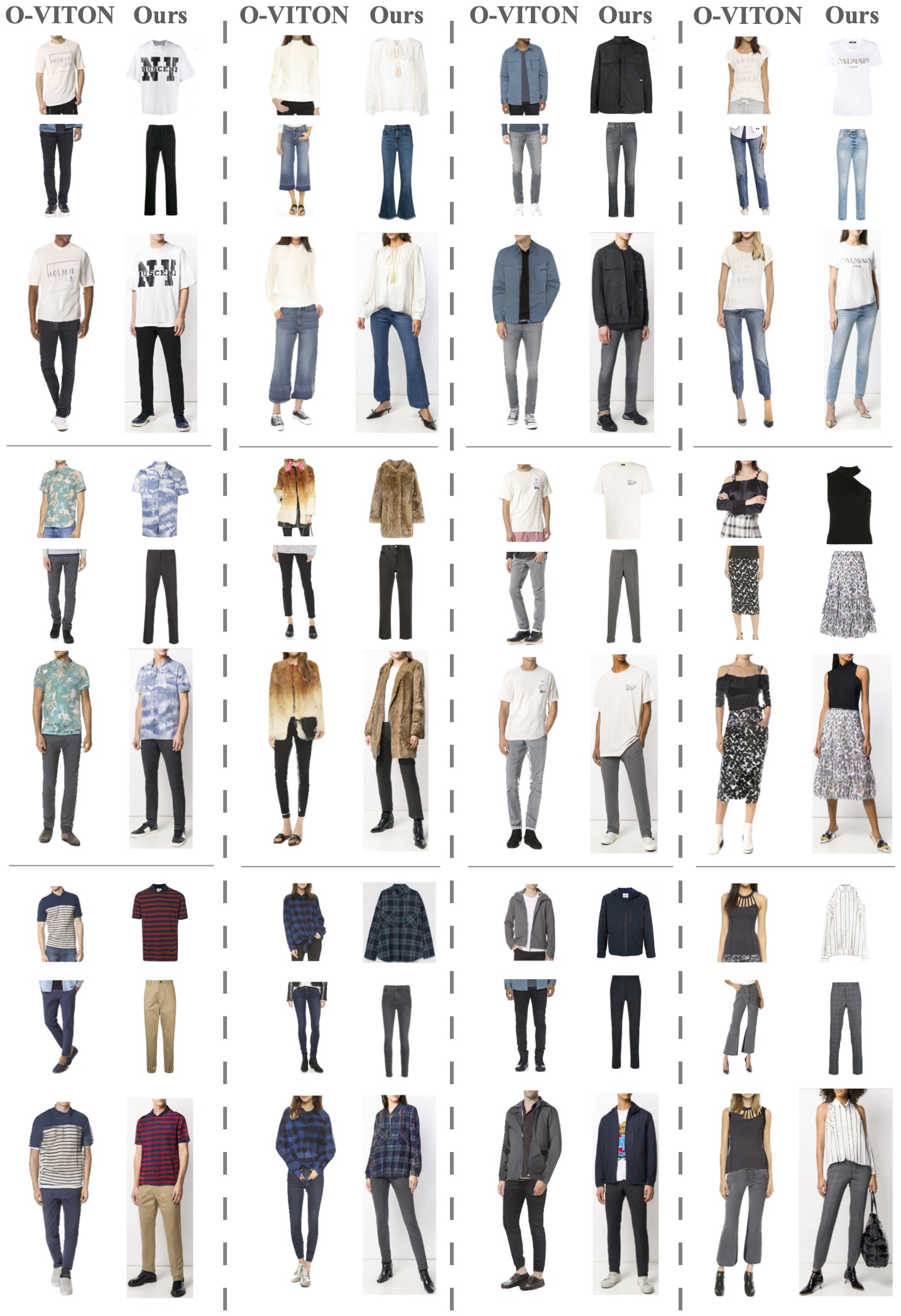}
\end{center}
           \vspace{-0.3cm}  
    \caption{Qualitative comparison with O-VITON~\cite{Neuberger_2020_CVPR}. The top two rows in each cell show the garments in the outfit and the bottom row in each cell shows generated try-on results.}  
          \label{fig:comparison}
\end{figure*}

We show more qualitative comparisons between our method and O-VITON~\cite{Neuberger_2020_CVPR}. Notice O-VITON~\cite{Neuberger_2020_CVPR} is the only prior work that supports multi-garment try-on, but they did not release their dataset or implementation. For a fair comparison, we found garment images that most closely resemble the garments chosen in~\cite{Neuberger_2020_CVPR} in terms of style, color, and texture. Image results for O-VITON are directly taken from their paper and supplementary materials. Notice the substantial improvement in generation quality in Figure~\ref{fig:comparison}.

 \begin{figure*}
\begin{center}
   \includegraphics[width=0.86\linewidth]{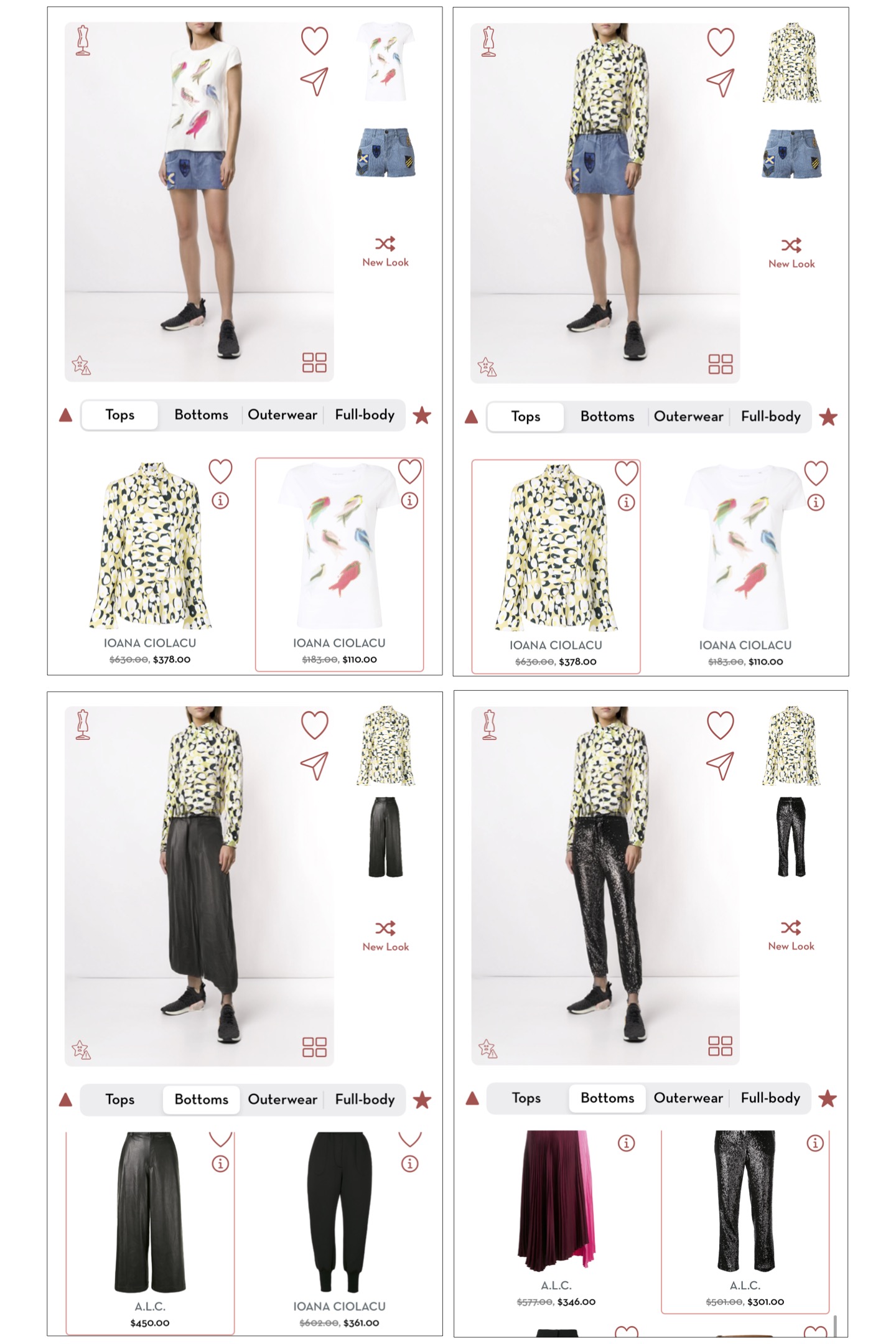}
\end{center}
           \vspace{-0.5cm}  
    \caption{ The figure shows examples of an interactive interface powered by our method. A user can select a garment listed below and our method will produce a visualization of the model wearing the selected garment in real-time. Garment images are products listed on an e-commerce site.}  
       \vspace{-0.5cm}  
\end{figure*}

 \begin{figure*}
\begin{center}
   \includegraphics[width=0.86\linewidth]{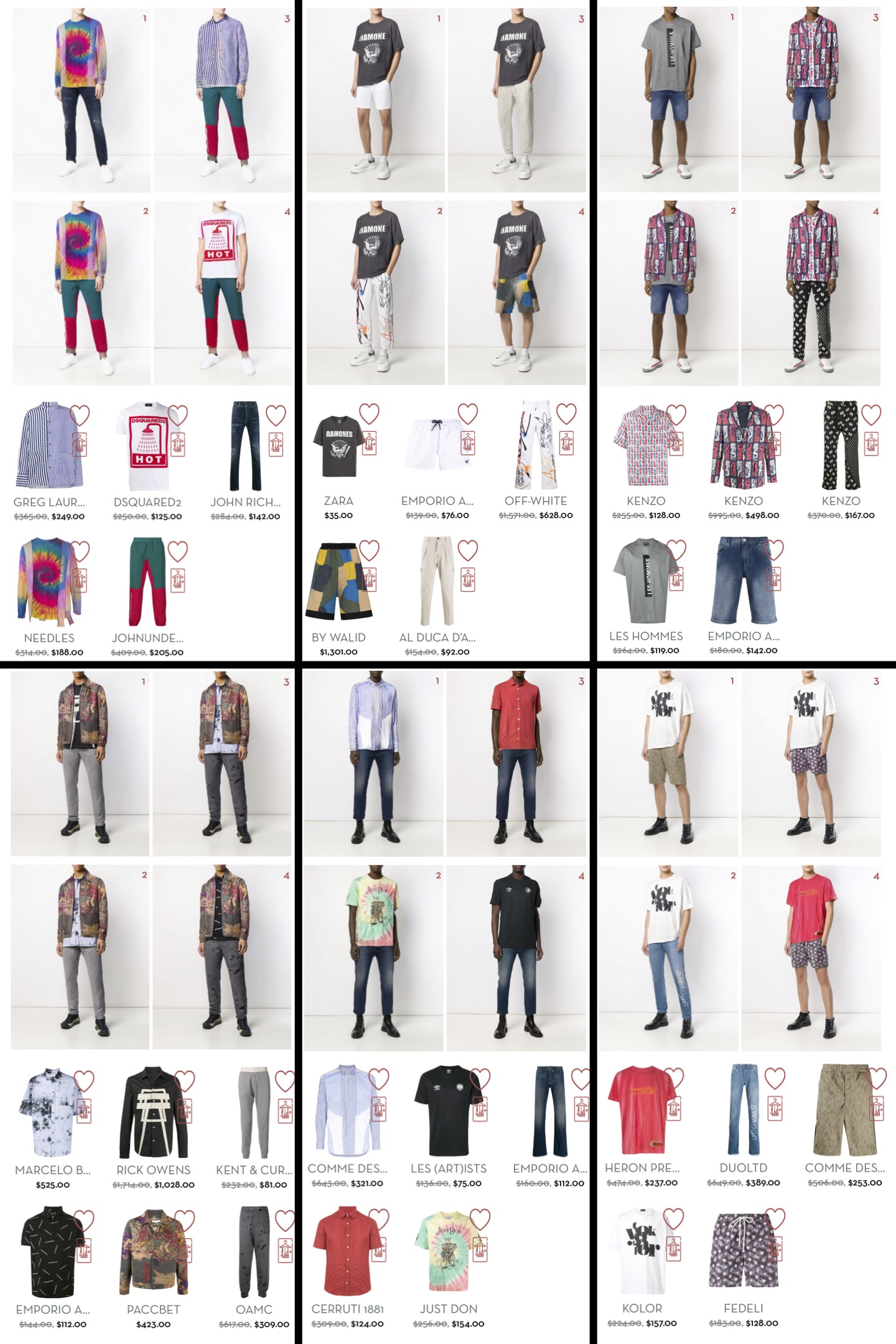}
\end{center}
           \vspace{-0.3cm}  
    \caption{ The figure shows outfits curated by users through the interactive interface (male models).}  
       \vspace{-0.3cm}  
          \label{fig:header}
\end{figure*}

 \begin{figure*}
\begin{center}
   \includegraphics[width=0.86\linewidth]{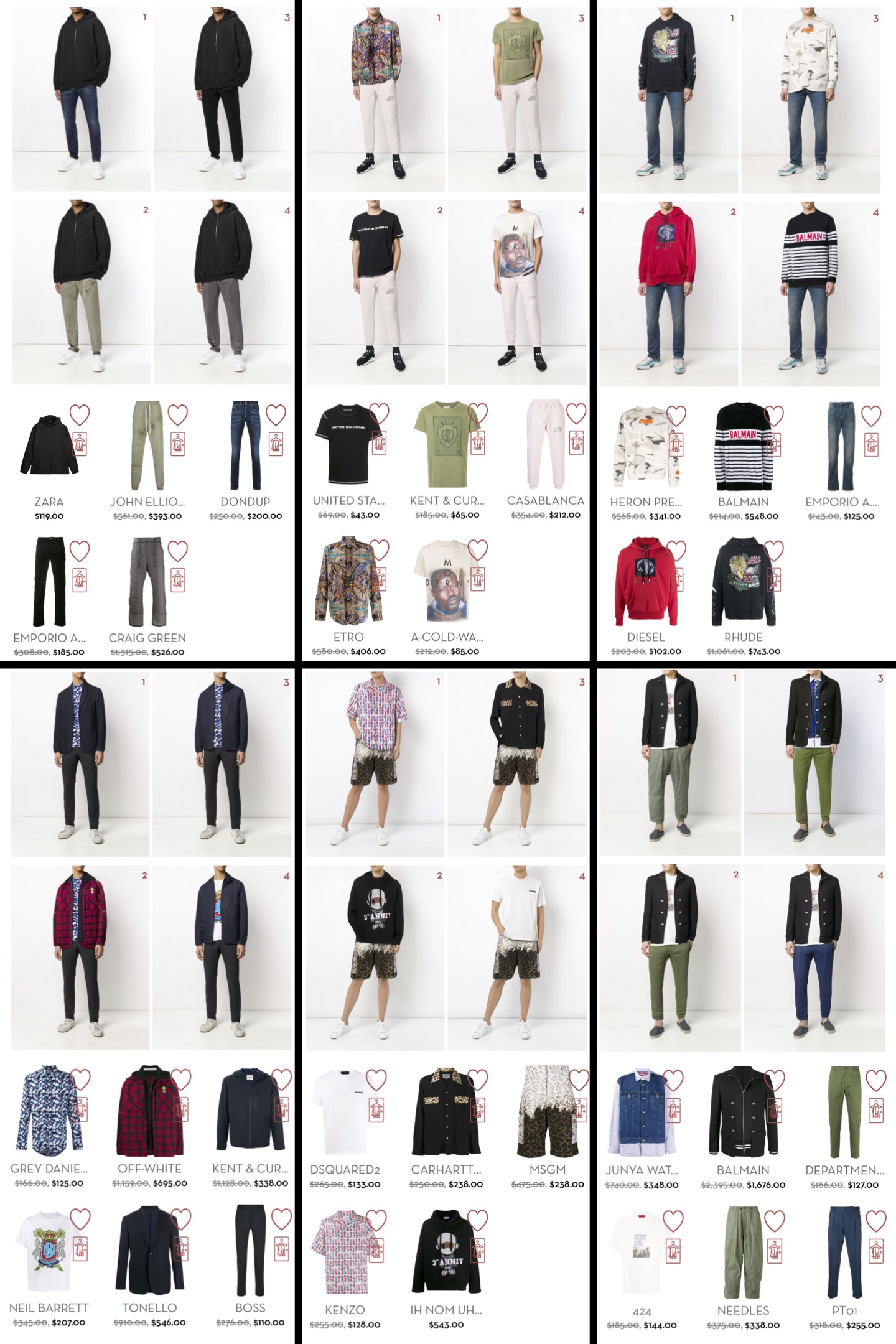}
\end{center}
           \vspace{-0.3cm}  
    \caption{ The figure shows outfits curated by users through the interactive interface (male models).}  
       \vspace{-0.3cm}  
          \label{fig:header}
\end{figure*}

 \begin{figure*}
\begin{center}
   \includegraphics[width=0.86\linewidth]{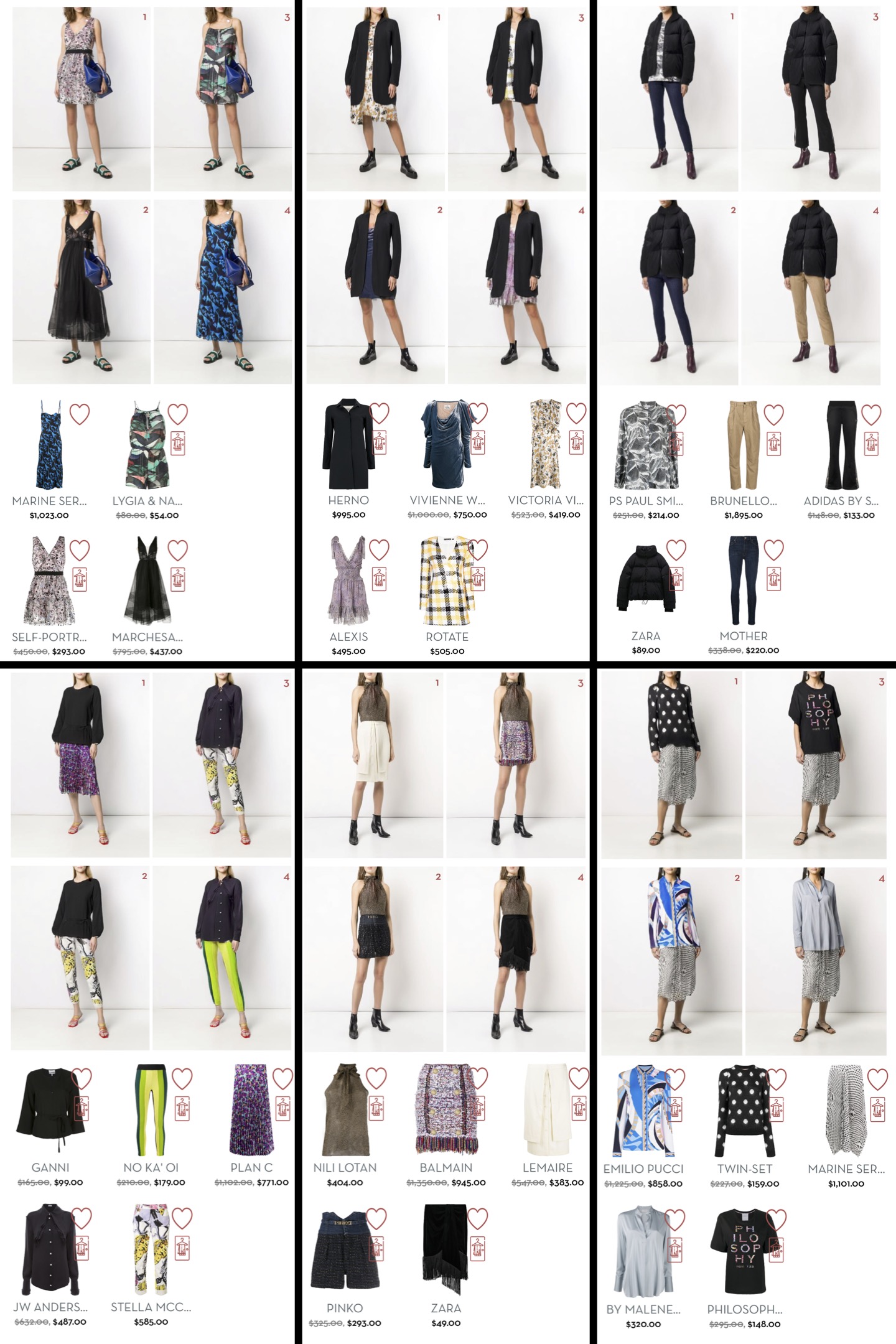}
\end{center}
           \vspace{-0.3cm}  
    \caption{ The figure shows outfits curated by users through the interactive interface (female models).}  
       \vspace{-0.3cm}  
          \label{fig:header}
\end{figure*}

 \begin{figure*}
\begin{center}
   \includegraphics[width=0.86\linewidth]{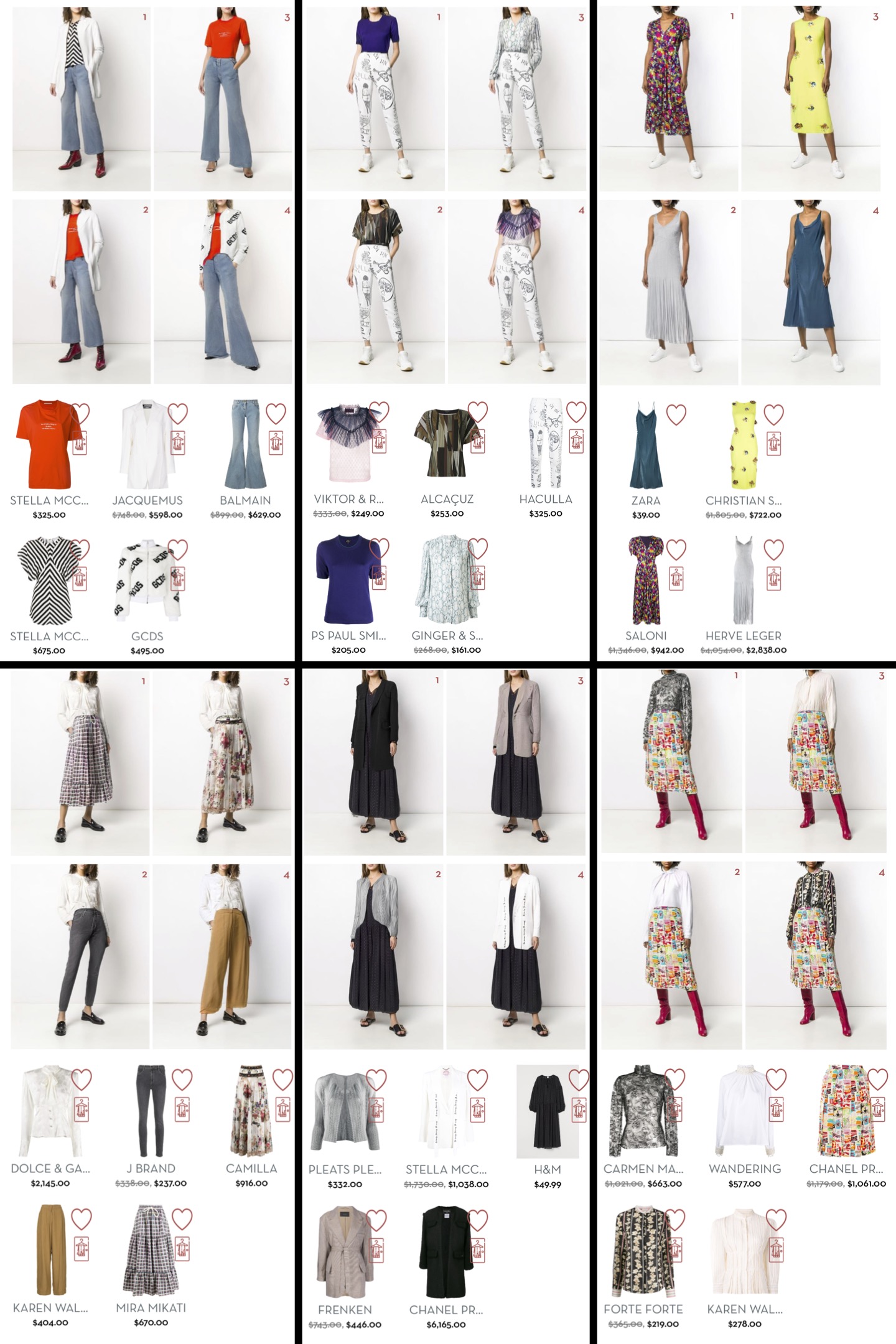}
\end{center}
           \vspace{-0.3cm}  
    \caption{ The figure shows outfits curated by users through the interactive interface (female models).}  
       \vspace{-0.3cm}  
          \label{fig:header}
\end{figure*}

\end{document}